%% file: main_jmlr.tex
\documentclass[twoside, 11pt]{article}

\usepackage{jmlr2e}

\let\proof\relax

\usepackage{amsmath, amsthm, amssymb, mathtools}
\usepackage[ansinew]{inputenc}
\usepackage{algorithm, algpseudocode} 

\usepackage[usenames]{xcolor}
\usepackage{bm, xspace}
\usepackage{pifont}
\usepackage{multirow, hhline, array, graphicx, multicol}
\usepackage{tikz, pgfplots}

\usepackage{fullpage}
\usepackage{makecell}
\newtheorem{Definition}{Definition}
\newtheorem{Theorem}{Theorem}
\newtheorem{Lemma}{Lemma}
\newtheorem{Claim}{Claim}

\newcommand{\ray}[1]{{\color{blue}\bf (#1)}}

\def\trainlets{\text{Trainlets}}

\firstpageno{1}

\begin{document}

\title{Provably Accurate Double-Sparse Coding}
\author{\name Thanh V. Nguyen \email thanhng@iastate.edu\\
 \addr Iowa State University, ECE Department
\AND \name Raymond K. W. Wong \email raywong@stat.tamu.edu\\
 \addr Texas A\&M University, Statistics Department 
\AND \name Chinmay Hegde
\thanks{This work is supported in part by the National Science Foundation under the grants CCF-1566281 and DMS-1612985. An abbreviated conference version will appear in the proceedings of AAAI 2018~\citep{nguyenAAAI}.}
\email chinmay@iastate.edu\\
\addr Iowa State University, ECE Department\\
}
\editor{TBD}

\maketitle

\input{math_commands}
\input{abstract}


\input{intro}
\input{setup}
\input{initialization}
\input{algorithm}
\input{empirical}
\input{conclusion}

\newpage
\input{appendix}

\vskip 0.2in
\bibliography{main}

\end{document}

%% file: math_commands.tex

\def\1{\bm{1}}

\newcommand*{\bigcdot}{\bullet}

\newcommand{\cA}[1]{A_{\bigcdot #1}}
\newcommand{\cAi}{\cA{i}}
\newcommand{\cAj}{\cA{j}}
\newcommand{\cAl}{\cA{l}}
\newcommand{\cAS}{\cA{S}}

\newcommand{\cB}[1]{A_{\bigcdot #1}}
\newcommand{\cBi}{\cB{i}}
\newcommand{\cBS}{\cB{S}}

\newcommand{\cD}[1]{D_{\bigcdot #1}}
\newcommand{\cDi}{\cD{i}}
\newcommand{\cDj}{\cD{j}}
\newcommand{\cDl}{\cD{l}}
\newcommand{\cDS}{\cD{S}}

\newcommand{\rA}[1]{A_{#1 \bigcdot}}
\newcommand{\rAj}{\rA{j}}
\newcommand{\rAR}{\rA{R}}
\newcommand{\rAl}{\rA{l}}

\newcommand{\AR}[1]{A_{R, #1}}

\newcommand{\cg}[1]{g_{\bigcdot #1}}
\newcommand{\cgi}{\cg{i}}
\newcommand{\cgS}{\cg{S}}
\newcommand{\gR}[1]{g_{R, #1}}

\def\ghat{\widehat{g}}
\def\Rhat{\widehat{R}}
\def\dhat{\widehat{d}}
\def\ehat{\widehat{e}}
\def\Muv{M_{u,v}}
\def\Mhuv{\widehat{M}_{u,v}}
\def\VarE{\mathcal{E}}
\def\Ne{N_{\varepsilon}}
\def\Rad{\mathcal{R}}
\def\Aupb{\mathcal{A}}
\def\Bupb{\mathcal{B}}

\newcommand*{\Asmp}[1]{Assumption \textnormal{\textbf{A#1}}}
\newcommand*{\AsmpB}[1]{Assumption \textnormal{\textbf{B#1}}}

\def\sgn{\textnormal{sgn}}
\def\supp{\textnormal{supp}}
\def\diag{\textnormal{diag}}
\def\card{\textnormal{card}}
\def\thres{\textnormal{threshold}}
\def\polylog{\textnormal{polylog}}
\def\iid{\text{i.i.d.}}
\def\whp{\text{w.h.p.}}
\def\wrt{\text{w.r.t.}}
\def\ie{\text{i.e.}}
\def\eg{\text{e.g.}}

\newcommand{\ceil}[1]{\left \lceil #1 \right \rceil}

\def\cmark{\ding{51}}
\def\xmark{\ding{55}}

\newcommand{\E}{\mathbb{E}}
\newcommand{\Prob}{\mathbb{P}}
\newcommand{\sigmae}{\sigma_\varepsilon}
\newcommand{\HardThres}{\mathcal{H}}

\newcommand{\R}{\mathbb{R}}

\DeclarePairedDelimiterX{\norm}[1]{\lVert}{\rVert}{#1} 
\DeclarePairedDelimiterX{\inprod}[2]{\langle}{\rangle}{#1, #2}
\DeclarePairedDelimiterX{\abs}[1]{\lvert}{\rvert}{#1}

\DeclarePairedDelimiterX{\bigO}[1]{(}{)}{#1}
\def\Otilde{\widetilde{O}}
\def\Omgtilde{\widetilde{\Omega}}

\newcommand{\sgnEvent}{\mathcal{F}_{x^*}}


%% file: abstract.tex
\begin{abstract}%
Sparse coding is a crucial subroutine in algorithms for various signal processing, deep learning, and other machine learning applications. The central goal is to learn an overcomplete dictionary that can sparsely represent a given input dataset. However, a key challenge is that storage, transmission, and processing of the learned dictionary can be untenably high if the data dimension is high. In this paper, we consider the double-sparsity model introduced by \cite{rubinstein10-sparsedl} where the dictionary itself is the product of a fixed, known basis and a data-adaptive sparse component. First, we introduce a simple algorithm for double-sparse coding that can be amenable to efficient implementation via neural architectures. Second, we theoretically analyze its performance and demonstrate asymptotic sample complexity and running time benefits over existing (provable) approaches for sparse coding. To our knowledge, our work introduces the first computationally efficient algorithm for double-sparse coding that enjoys rigorous statistical guarantees. Finally, we support our analysis via several numerical experiments on simulated data, confirming that our method can indeed be useful in problem sizes encountered in practical applications.
\end{abstract}

%% file: intro.tex
\section{Introduction}
\label{intro}

\subsection{Motivation}

Representing signals as sparse linear combinations
of atoms from a dictionary is a popular approach
in many domains.
In this paper, we study the problem of
\textit{dictionary learning} (also known as sparse coding),
where the goal is to learn an efficient basis (dictionary) that
 represents the underlying class of signals well.
In the typical sparse coding setup, the dictionary is \emph{overcomplete} (\ie, the cardinality of the dictionary exceeds the ambient signal dimension) while the representation is \emph{sparse} (\ie, each signal is encoded by a combination of only very few dictionary atoms.)

Sparse coding has a rich history in diverse fields such as signal processing, machine learning, and computational neuroscience. 
Discovering optimal basis representations of data is a central focus of image analysis~\citep{donoho,elad06-denoising,elad2}, and dictionary learning has proven widely successful in imaging problems such as denoising, deconvolution, inpainting, and compressive sensing~\citep{elad06-denoising, candes05-decoding, elad2}.
Sparse coding approaches have also been used as a core building block of deep learning systems for prediction~\citep{lista,boureau2010learning} and associative memory~\citep{mazumdar2017}.  
Interestingly, the seminal work by \citet{olshausen97-sc} has shown intimate connections between sparse coding and neuroscience: the dictionaries learned from image patches of natural scenes bear strikingly resemblance to spatial receptive fields observed in mammalian primary visual cortex.

From a mathematical standpoint, the sparse coding problem is formulated as follows.
Given $p$ data samples $Y = [y^{(1)}, y^{(2)}, \dots, y^{(p)}] \in\R^{n\times p}$, the goal is to find a dictionary $D \in \R^{n\times m}$ ($m > n$) and corresponding sparse code
vectors $X = [x^{(1)}, x^{(2)}, \dots, x^{(p)}]\in\R^{m\times p}$
such that the representation $D X$ fits the data samples as well as possible.
Typically, one obtains the dictionary and the code vectors
as the solution to the following optimization problem:
\begin{align}
  \begin{split}
\min_{D, X}\mathcal{L}(D, X) &= \frac{1}{2}\sum^p_{j=1}\Vert  y^{(j)} - Dx^{(j)}\Vert^2_2,\label{eq_objective_func} \\
\text{s.t.}~&\sum^p_{j=1}\mathcal{S}(x^{(j)}) \leq S
\end{split}
\end{align}
where $\mathcal{S}(\cdot)$ is some sparsity-inducing penalty function on the code vectors, such as the $\ell_1$-norm.
The objective function $\mathcal{L}$ controls the reconstruction error
while the constraint enforces the sparsity of the representation.

However, even a cursory attempt at solving the optimization problem \eqref{eq_objective_func} reveals the following obstacles:

\begin{enumerate}

\item {\itshape{\textbf{Theoretical challenges}}}. The constrained optimization problem \eqref{eq_objective_func} involves a non-convex (in fact, bilinear) objective function, as well as 
potentially non-convex constraints depending on the choice of the sparsity-promoting function $\mathcal{S}$ (for example, the $\ell_0$ function.)
Hence, obtaining \emph{provably} correct algorithms for this problem can be challenging. Indeed, the vast majority of practical approaches for sparse coding have been heuristics~\citep{engan99-mod, aharon06-ksvd, mairal09-online}; 
Recent works in the theoretical machine learning community have bucked this trend, providing provably accurate algorithms if certain assumptions are satisfied~\citep{spielman12-exact,agarwal2014learning, arora15-neural, sun2015complete, blasiok16-erspud-improved, law16_note_erspud, bartlett17}. However, relatively few of these newer methods have been shown to provide good empirical performance in actual sparse coding problems.

\item {\itshape{\textbf{Practical challenges}}}. Even if theoretical correctness issues were to be set aside, and we are somehow able to efficiently learn sparse codes of the input data, we often find that applications using such learned sparse codes encounter \emph{memory} and \emph{running-time} issues. Indeed, in the overcomplete case, only the storage of the learned dictionary $D$ incurs $mn = \Omega(n^2)$ memory cost, which is prohibitive when $n$ is large. Therefore, in practical applications (such as image analysis) one typically resorts to chop the data into smaller blocks (\eg, partitioning image data into patches) to make the problem manageable. 

\end{enumerate}

A related line of research has been devoted to learning dictionaries that obey some type
of \emph{structure}. Such structural information can be leveraged to incorporate prior knowledge of underlying signals as well as to resolve computational challenges due to the data dimension.
For instance, the dictionary is assumed to be separable, or obey a convolutional structure. 
One such variant is the \emph{double-sparse} coding problem~\citep{rubinstein10-sparsedl, sulam16-trainlets} where the dictionary $D$ \emph{itself} exhibits a sparse structure. To be specific, the dictionary is expressed as:
$$D = \Phi A,$$ 
\ie, it is composed of a known ``base dictionary" $\Phi \in \R^{n
  \times n}$, and a learned ``synthesis'' matrix $A \in
\R^{n \times m}$ whose columns are sparse. The base dictionary $\Phi$ is typically any
fixed basis chosen
according to domain knowledge, while the synthesis matrix $A$ is
column-wise sparse and is to be learned from the data. The basis $\Phi$ is typically orthonormal (such as the canonical or wavelet basis); however, there are cases where the base
dictionary $\Phi$ is overcomplete~\citep{rubinstein10-sparsedl, sulam16-trainlets}. 

There are several reasons why such the double-sparsity model can be useful. First, the double-sparsity assumption is rather appealing from a conceptual standpoint, since it lets us combine the knowledge of decades of modeling efforts in harmonic analysis with the flexibility of learning new representations tailored to specific data families. Moreover, such a double-sparsity model has computational benefits. If the columns of $A$ are (say) $r$-sparse (i.e., each column contains no more than $r \ll n$ non-zeroes) then the overall burden of storing, transmitting, and computing with $A$ is much lower than that for general unstructured dictionaries. Finally, such a model lends itself well to \emph{interpretable} learned features if the atoms of the base dictionary are semantically meaningful.

All the above reasons have spurred researchers to develop a series of algorithms to learn doubly-sparse codes~\citep{rubinstein10-sparsedl,sulam16-trainlets}. However, despite their empirical promise, no theoretical analysis of their performance have been reported in the literature and to date, we are unaware of a provably accurate, polynomial-time algorithm for the double-sparse coding problem. Our goal in this paper is precisely to fill this gap.

\subsection{Our Contributions}

In this paper, we provide a new framework for double-sparse coding. To the best of our knowledge, our approach is the first method that enjoys \emph{provable} statistical and algorithmic guarantees for this problem. In addition, our approach enjoys three benefits: we demonstrate that the method is \emph{neurally plausible} (\ie, its execution can plausibly be achieved using a neural network architecture), 
\emph{robust} to noise, as well as \emph{practically useful}. 

Inspired by the aforementioned recent theoretical advances in sparse coding, 
we assume a learning-theoretic setup where the data samples arise from a ground-truth {generative model}. Informally, suppose there exists a true (but unknown) synthesis matrix $A^*$ that is column-wise $r$-sparse, and the $i^{\textrm{th}}$ data sample is generated as:
$$
y^{(i)} = \Phi A^* x^{*(i)} +~\text{noise},~~~i=1,2,\ldots,p
$$
where the code vector $x^{*(i)}$ is independently drawn from a distribution supported on the set of $k$-sparse vectors. We desire to learn the underlying matrix $A^*$. Informally, suppose that the synthesis matrix $A^*$ is \emph{incoherent} (the columns of $A^*$ are sufficiently close to orthogonal) and has bounded spectral norm. Finally, suppose that the number of dictionary elements, $m$, is at most a constant multiple of $n$. All of these assumptions are standard\footnote{We clarify both the data and the noise model more concretely in Section~\ref{setup} below.}.

We will demonstrate that 
the true synthesis matrix $A^*$
can be recovered (with small error) in a tractable manner as sufficiently many samples are provided. Specifically, we make the following novel contributions:

\begin{enumerate}

\item We propose a new algorithm that produces a coarse estimate of the synthesis matrix that is sufficiently close to the ground truth $A^*$. In contrast with previous double-sparse coding methods (such as~\citet{sulam16-trainlets}), our algorithm is \emph{not} based on alternating minimization. Rather, it builds upon spectral initialization-based ideas that have recently gained popularity in non-convex machine learning~\citep{zhang2016spectral,wang2016unified}.

\item Given the above coarse estimate of the synthesis matrix $A^*$, we propose a descent-style algorithm to refine the above estimate of $A^*$. This algorithm is simpler than previously studied double-sparse coding algorithms (such as the Trainlets approach of~\citet{sulam16-trainlets}), while still giving good 
statistical performance. 
Moreover, this algorithm can be realized in a manner amenable to neural implementations.

\item We provide a rigorous analysis of both algorithms. Put together, our analysis produces the first provably polynomial-time algorithm for
double-sparse coding. We show that the algorithm provably returns a good estimate of the ground-truth; in particular, in the absence of noise we prove that $\Omega(mr~\polylog~n)$ samples are sufficient for a good enough initialization in the first algorithm, as well as guaranteed linear convergence of the descent phase up to a precise error parameter that can be interpreted as the radius of convergence. 

Indeed, our analysis shows that employing the double-sparsity model helps in this context, and leads to a strict improvement in sample complexity, as well as running time over previous rigorous methods for (regular) sparse coding such as~\citet{arora15-neural}.

\item We also analyze our approach in a more realistic setting with the presence of additive noise and demonstrate its stability. 
We prove
that $\Omega(mr~\polylog~n)$ samples are sufficient to obtain a
good enough estimate in the initialization, and also to obtain
guaranteed linear convergence during descent to provably recover
$A^*$. 

\item We underline the benefit of the double-sparse structure over the regular model by analyzing the algorithms in~\citet{arora15-neural} under the noisy setting. As a result, we obtain the sample complexity $O\bigl( (mk + \sigmae^2\frac{mn^2}{k})\polylog~n\bigr)$, which demonstrates a negative effect of noise on this approach. 

\item We rigorously develop a hard thresholding intialization that extends the spectral scheme in~\citet{arora15-neural}. Additionally, we provide more results for the case where $A$ is orthonormal, sparse dictionary to relax the condition on $r$, which may be of independent interest.

\item While our analysis mainly consists of sufficiency results and involves several (absolute) unspecified constants, in practice we have found that these constants are reasonable. We justify our observations by reporting a suite of numerical experiments on synthetic test datasets.  

\end{enumerate}

\begin{table}[]
  \scriptsize
  \centering
\begin{tabular}{|c|c|c|c|c|c|}
	\hline
	{Setting} & Reference & \makecell{Sample complexity \\ (w/o noise)} & \makecell{Sample complexity \\ (w/ noise)} & \makecell{Upper bound on \\  running time} & {Expt} \\
	\hline \hline
	\multirow{6}{*}{Regular} & MOD \citep{engan99-mod} & \xmark & \xmark & \xmark & \cmark \\
	\hhline{~-----}
	& K-SVD \citep{aharon06-ksvd} & \xmark & \xmark & \xmark & \cmark \\
	\hhline{~-----}
	& \citet{spielman12-exact} & $O(n^2 \log n)$ & \xmark & $\Omgtilde(n^4)$ & \cmark \\
	\hhline{~-----}
	& \citet{arora14-new-algorithms} & $\Otilde(m^2/k^2)$ & \xmark &  $\Otilde(np^2)$ & \xmark \\
	\hhline{~-----}
	& \citet{gribonval15-sparsespurious} & $O(nm^3)$ & $O(nm^3)$ & \xmark & \xmark \\
	\hhline{~-----}
	& \citet{arora15-neural} &  $\Otilde(mk)$ & \xmark & $\Otilde(mn^2p)$ & \xmark \\
	\hline \hline
	\multirow{4}{*}{\makecell{Double \\ Sparse}} & Double Sparsity \citep{rubinstein10-sparsedl} & \xmark & \xmark & \xmark & \cmark \\
	\hhline{~-----}
	& \citet{gribonval15-sample} & $\Otilde(mr)$  & $\Otilde(mr)$ 
                                                                                                                         & \xmark & \xmark \\
	\hhline{~-----}
	& Trainlets \citep{sulam16-trainlets} & \xmark & \xmark & \xmark
                                                                                                                                                             & \cmark \\
	\hhline{~-----}
	& This paper & $\Otilde(mr)$ & $\Otilde(mr + \sigmae^2\frac{mnr}{k})$ & $\Otilde(mnp)$ & \cmark \\
	\hline
\end{tabular}
  \caption{\small \sl Comparison of various sparse coding techniques.
     Expt: whether
    numerical experiments have been conducted. \xmark\, in all other columns indicates no provable guarantees.
    Here, $n$ is the signal
    dimension, and $m$ is the number of atoms. The sparsity levels for $A$ and
    $x$ are $r$ and $k$ respectively, and $p$ is the sample size.
    \label{tbl_overview}}
\end{table}



Overall, our approach results in strict improvement in sample
complexity, as well as running time, over previous rigorously analyzed
methods for (regular) sparse coding, such as~\citet{arora15-neural}.
See Table~\ref{tbl_overview} for a detailed comparison. 

\subsection{Techniques}

At a high level, our method is an adaptation of the seminal approach of~\citet{arora15-neural}. As is common in the statistical learning literature, we assume a ``ground-truth" generative model for the observed data samples, and attempt to estimate the parameters of the generative model given a sufficient number of samples. In our case, the parameters correspond to the synthesis matrix $A^*$, which is column-wise $r$-sparse. The natural approach is to formulate a loss function in terms of $A$ such as Equation~\eqref{eq_objective_func}, and perform gradient descent with respect to the surface of the loss function to learn $A^*$.

The key challenge in sparse coding is that the gradient is inherently coupled with the \emph{codes} of the training samples (\ie, the columns of $X^*$), which are unknown \emph{a priori}. However, the main insight of~\cite{arora15-neural} is that within a small enough neighborhood of $A^*$, a noisy version of $X^*$ can be estimated, and therefore the overall method is similar to performing \emph{approximate gradient descent.}
Formulating the actual algorithm as a noisy variant of approximate gradient descent allows us to overcome the finite-sample variability of the loss, and obtain a descent property
directly related to (the population parameter) $A^*$.

The second stage of our approach (\ie, our descent-style algorithm) leverages this intuition. However, instead of standard gradient descent, we perform approximate \emph{projected} gradient descent, such that the column-wise $r$-sparsity property is enforced in each new estimate of $A^*$. 
Indeed, such an extra projection step is critical in showing a sample complexity improvement over the existing approach of~\citet{arora15-neural}.
The key novelty is in figuring out how to perform the projection in each gradient iteration. For this purpose, we develop a novel initialization algorithm that identifies the locations of the non-zeroes in $A^*$ even before commencing the descent phase. This is nontrivially different from initialization schemes used in previous rigorous methods for sparse coding, and the analysis is somewhat more involved.

In~\citet{arora15-neural},
(the principal eigenvector of) a weighted covariance matrix of $y$
(estimated by the weighted average of outer products $y_iy_i^T$)
is shown to provide a coarse estimate of a dictionary atom.
We extend this idea and rigoriously show that the
diagonal of the weighted covariance matrix serves as a good indicator
of the support of a column in $A^*$.
The success relies on the concentration of
the diagonal vector with dimension $n$, instead of the covariance matrix
with dimensions $n\times n$. With the support selected, our scheme only utilizes a reduced weighted covariance matrix with dimensions at most $r\times r$.
This initialization scheme enables us to effectively reduce the dimension of the problem, and therefore leads to significant improvement in sample complexity and running time over previous (provable) sparse coding methods when the data representation sparsity $k$ is much smaller than $m$.

Further, we rigorously analyze the proposed algorithms in the presence of noise with a bounded expected norm. Our analysis shows that our method is stable, and in the case of i.i.d. Gaussian noise with bounded expected $\ell_2$-norms, is at least a polynomial factor better than previous polynomial time algorithms for sparse coding.

The empirical performance of our proposed method is demonstrated
by a suite of numerical experiments on synthetic datasets.
In particular, we show that our proposed methods are simple and practical, and improve upon previous provable algorithms for sparse coding. 


  

\subsection{Paper Organization}

The remainder of this paper is organized as
follows. Section~\ref{setup} introduces notation, key model
assumptions, and informal statements of our main theoretical
results. Section~\ref{init} outlines our initialization algorithm (along with supporting theoretical results)
while Section~\ref{descent} presents our descent algorithm (along with supporting theoretical results).
Section~\ref{empirical} provides a numerical study of the efficiency of our proposed algorithms, and compares it with previously proposed methods. Finally, Section~\ref{conc}
concludes with a short discussion. All technical proofs are relegated to the appendix. 


%% file: setup.tex
\section{Setup and Main Results}
\label{setup}

\subsection{Notation}
\label{notation}

We define $[m] \triangleq \{1, \ldots, m\}$ for any integer
$m > 1$. For any vector $x = [x_1, x_2, \ldots, x_m]^T \in\R^{m}$,
we write $\supp(x)\triangleq \{i \in [m]: x_i \neq 0 \}$ as the support set of $x$.
Given any subset $S \subseteq [m]$, $x_S$ corresponds to the sub-vector of $x$ indexed by the elements of $S$. For any matrix $A \in \R^{n\times m}$, we use $\cAi$
and $\rAj^T$ to represent the $i$-th column and the $j$-th row respectively. For some appropriate sets $R$ and $S$,
let $\rAR$ (respectively, $\cAS$) be the submatrix of $A$ with rows
(respectively columns) indexed by the elements in $R$ (respectively $S$).
In addition,
for the $i$-th column $\cAi$, we use
$A_{R, i}$ to denote the sub-vector indexed by the elements of $R$.
For notational simplicity, we use
$\rAR^T$ to indicate $(\rAR)^T$, the tranpose of $A$ after
a row selection.
Besides, we use $\circ$ and $\sgn(\cdot)$ to represent the element-wise Hadamard operator
and the element-wise sign function respectively. Further, $\mathrm{threshold}_{K}(x)$
is a thresholding operator that replaces any elements of $x$ with magnitude less than $K$ by zero.

The $\ell_2$-norm $\norm{x}$ for a vector $x$ and the spectral norm $\norm{A}$ for a matrix $A$
appear several times.
In some cases,
we also utilize the Frobenius norm
$\norm{A}_F$ and
the operator norm
$\norm{A}_{1,2}\triangleq\max_{\norm{x}_1 \leq 1} \norm{Ax}$.
The norm $\norm{A}_{1, 2}$ is essentially the maximal Euclidean norm of any column of $A$.

For clarity, 
we adopt asymptotic notations extensively.
We write $f(n) = O(g(n))$ (or $f(n) = \Omega(g(n))$) if
$f(n)$ is upper bounded (respectively, lower bounded) by $g(n)$ up
to some positive constant. Next, $f(n) = \Theta(g(n))$ if and only if $f(n) = O(g(n))$ and $f(n) = \Omega(g(n))$.
Also $\Omgtilde$ and $\Otilde$ represent $\Omega$ and $O$  up to
a multiplicative poly-logarithmic factor respectively. Finally $f(n) = o(g(n))$ (or $f(n) = \omega(g(n))$) if
$\lim_{n\rightarrow\infty} |f(n)/g(n)|=0$ ($\lim_{n\rightarrow\infty}
|f(n)/g(n)|=\infty$).

Throughout the paper, we use the phrase ``with high probability'' (abbreviated to \whp) to describe an event with failure probability of order at most $n^{-\omega(1)}$. In addition, $g(n)=O^*(f(n))$ means $g(n)\le Kf(n)$
for some small enough constant $K$.

\subsection{Model}
\label{model}

Suppose that the observed samples are given by
$$y^{(i)} = D x^{*(i)} + \varepsilon,~~i = 1, \ldots, p,$$ 
\ie, we are given $p$ samples of $y$ generated from a fixed (but unknown) dictionary $D$ where the sparse code $x^*$ and the error $\varepsilon$ are drawn from a joint distribution $\mathcal{D}$ specified below. In the double-sparse setting, the dictionary is assumed to follow a
decomposition $D = \Phi A^*$, where $\Phi\in\R^{n\times n}$ is a known
\emph{orthonormal} basis matrix and $A^*$ is an unknown, ground truth
synthesis matrix. An alternative (and interesting) setting is an overcomplete $\Phi$
with a square $A^*$, which our analysis below does not cover; we defer this to future work.
Our approach relies upon the following
assumptions on the synthesis dictionary $A^*$:

\begin{enumerate}
\item[\textbf{A1}] $A^*$ is overcomplete (i.e.,~$m \geq n$)
  with $m = O(n)$.
\item[\textbf{A2}] $A^*$ is $\mu$-incoherent, \ie, for all $i \neq j$, $\abs{\inprod{\cA{i}^*}{\cA{j}^*}} \leq \mu/\sqrt{n}$.
\item[\textbf{A3}] $\cAi^*$ has at most $r$ non-zero elements, and is normalized such that $\norm{\cAi^*} = 1$ for all $i$. Moreover, $\abs{A^*_{ij}} \geq \tau$ for $A^*_{ij} \neq 0$ and $\tau = \Omega(1/\sqrt{r})$.
\item[\textbf{A4}] $A^*$ has bounded spectral norm such that $\norm{A^*} \leq O(\sqrt{m/n})$.
\end{enumerate}
All these assumptions are standard. In \Asmp{2}, the incoherence $\mu$ is
typically of order $O(\log n)$ with high probability for a normal random
matrix~\citep{arora14-new-algorithms}. Assumption
\textbf{A3} is a common assumption in sparse signal recovery. The bounded spectral norm assumption is also standard~\citep{arora15-neural}. 
%
In addition to Assumptions \textbf{A1-A4}, we make the following distributional
assumptions on $\mathcal{D}$:
\begin{enumerate}
\item[\textbf{B1}] Support $S=\supp(x^*)$ is of size at most $k$ and uniformly drawn without replacement from $[m]$ such that $\Prob[i \in S] = \Theta(k/m)$ and  $\Prob[i, j \in S] = \Theta(k^2/m^2)$ for some $i, j \in [m]$ and $i \neq j$.
\item[\textbf{B2}] The nonzero entries $x^*_S$ are pairwise independent and sub-Gaussian given the support $S$ with $\E[x^*_i|i \in S] = 0$ and $\E[x^{*2}_i|i \in S] = 1$.
\item[\textbf{B3}] For $i \in S$, $|{x^*_i}| \geq C$ where $0 < C \leq 1$. 
\item[\textbf{B4}] The additive noise $\varepsilon$ has i.i.d.\ Gaussian entries with variance $\sigmae^2$ with $\sigmae = O(1/\sqrt{n})$.
\end{enumerate}
%

For the rest of the paper, we set
$\Phi = I_n$, the identity matrix of size $n$. This only
simplifies the arguments but does not change the problem because one can study
an equivalent model: 
$$y' = A x^* + \varepsilon',$$
where $y'=\Phi^T y$ and $\varepsilon' = \Phi^T \varepsilon$, as $\Phi^T\Phi=I_n$.
Due to the Gaussianity of $\varepsilon$, $\varepsilon'$ also has independent
entries. Although this property is specific to Gaussian noise, all the analysis carried out below can be extended to sub-Gaussian noise with minor (but rather tedious) changes in concentration arguments.

Our goal is to devise an algorithm that produces an provably ``good'' estimate of $A^*$. For this, we need to define a suitable measure of ``goodness''. 
We use the following notion of distance 
that measures the maximal column-wise difference in $\ell_2$-norm under some suitable transformation.
\begin{Definition}[$(\delta, \kappa)$-nearness]
  $A$ is said to be $\delta$-close to $A^*$ if there is a permutation $\pi : [m] \rightarrow [m]$ and a sign flip $\sigma : [m] : \{\pm 1\}$ such that $\norm{\sigma(i) \cA{\pi(i)} - \cAi^*} \leq \delta$ for every $i$. In addition, $A$ is said to be $(\delta, \kappa)$-near to $A^*$ if $\norm{\cA{\pi} - A^*} \leq \kappa \norm{A^*}$ also holds.
\label{def_closeness}
\end{Definition}
For notational simplicity, in our theorems we simply replace $\pi$ and $\sigma$ in Definition \ref{def_closeness} with the identity permutation $\pi(i) = i$ and the positive sign $\sigma(\cdot) = +1$ while keeping in mind that in reality we are referring to one element of the equivalence class of all permutations and sign flip transforms of $A^*$.

We will also need some technical tools from~\citet{arora15-neural} to analyze our gradient descent-style method. Consider any iterative algorithm that looks for a desired solution $z^* \in \R^n$ to optimize some function $f(z)$. Suppose that the algorithm produces a sequence of estimates $z^1, \dots, z^s$ via the update rule:
$$z^{s+1} = z^s - \eta g^s,$$ for some vector $g^s$ and scalar step size $\eta$. The goal is to characterize ``good'' directions $g^s$ such that the sequence converges to $z^*$ under the Euclidean distance. The following gives one such sufficient condition for $g^s$.

\begin{Definition}
  \label{def_correlated_direction}
  A vector $g^s$ at the $s^{\textrm{th}}$ iteration is $(\alpha, \beta,
  \gamma_s)$-correlated with a desired solution $z^*$ if
  $$\inprod{g^s} {z^s - z^*} \geq \alpha \norm{z^s - z^*}^2 + \beta \norm{g^s}^2 - \gamma_s.$$
\end{Definition}

We know from convex optimization that if $f$ is $2\alpha$-strongly convex and $1/2\beta$-smooth, and $g^s$ is chosen as the gradient $\nabla_z f(z)$, then $g^s$ is $(\alpha, \beta, 0)$-correlated with $z^*$.  In our setting, the desired solution corresponds to $A^*$, the ground-truth synthesis matrix. 
In \cite{arora15-neural}, it is shown that $g^s= \E_y[(A^s x-y)\sgn(x)^T]$,
where $x=\thres_{C/2}((A^{s})^Ty)$ indeed satisfies Definition~\ref{def_correlated_direction}.
This $g^s$ is a population quantity and not explicitly available, but
one can estimate such $g^s$ using an empirical average. The corresponding estimator $\ghat^s$ is a random variable, 
so we also need
a related \emph{correlated-with-high-probability} condition: 

\begin{Definition} 
  \label{def_correlated_direction_whp}
  A direction $\ghat^s$ at the $s^{\textrm{th}}$ iteration is $(\alpha, \beta,
  \gamma_s)$-correlated-w.h.p. with a desired solution $z^*$ if, w.h.p.,
  $$\inprod{\ghat^s} {z^s - z^*} \geq \alpha \norm{z^s - z^*}^2 + \beta  \norm{\ghat^s}^2 - \gamma_s.$$

\end{Definition}
From Definition \ref{def_correlated_direction}, one can establish a form of descent property
in each update step, as shown in Theorem \ref{thm_descent_from_correlation_z}.

\begin{Theorem} 
   \label{thm_descent_from_correlation_z}
  Suppose that $g^s$ satisfies the condition described in Definition \ref{def_correlated_direction} for $s =
  1, 2, \dots, T$. Moreover, $0 < \eta \leq 2\beta$ and $\gamma =
  \max_{s=1}^T\gamma_s$. Then, the following holds for all $s$:
  $$\norm{z^{s+1} - z^*}^2 \leq (1-2\alpha \eta) \norm{z^s - z^*}^2 + 2\eta\gamma_s.$$
In particular, the above update converges geometrically to $z^*$ with an error $\gamma/\alpha$. That is,
  $$\norm{z^{s+1} - z^*}^2 \leq (1-2\alpha \eta)^s \norm{z^0 - z^*}^2 + 2\gamma/\alpha.$$
\end{Theorem}
We can obtain a similar result for Definition \ref{def_correlated_direction_whp} except that
$\norm{z^{s+1} - z^*}^2$ is replaced with its expectation. 

Armed with the above tools, we now state some informal versions of our main results:

\begin{Theorem}[Provably correct initialization, informal]
  \label{result_thm_neural_initialization}
  There exists a neurally plausible algorithm to produce an initial estimate $A^0$ that
  has the correct 
  support and is $(\delta, 2)$-near to $A^*$ with high probability. Its running time and sample complexity are $\Otilde(mnp)$
  and $\Otilde(mr)$ respectively.
  This algorithm works when the sparsity level satisfies $r = O^*(\log n)$. 
\end{Theorem}
Our algorithm can be regarded as an extension of~\citet{arora15-neural} to the
double-sparse setting. 
 It reconstructs the support of one single column and then estimates its direction in the subspace defined by the support. Our proposed algorithm enjoys neural plausibility by implementing a thresholding non-linearity and Oja's update rule. We provide a neural implementation of our algorithm in Appendix~\ref{neural_implementation}. 
 The adaption to the sparse
structure results in a strict improvement upon the original algorithm
both in running time and sample complexity. However, our algorithm is
limited to the sparsity level $r = O^*(\log n)$, which is rather small
but plausible from
the modeling standpoint. 
For comparison, we analyze a natural extension
of the algorithm of~\citet{arora15-neural} with an extra hard-thresholding step for
every learned atom.
We obtain the same order restriction on $r$,
but somewhat worse bounds on sample complexity and running time.
The details are found in Appendix~\ref{appdx_improve_arora}.

We hypothesize that a stronger
incoherence assumption can lead to provably correct initialization for a much wider range of $r$.
For purposes of theoretical analysis, we consider the special case of a \emph{perfectly incoherent} synthesis matrix $A^*$ such that $\mu = 0$ and $m = n$. In this case, we can indeed improve the sparsity parameter to $r = O^*\bigl( \min(\frac{\sqrt{n}}{\log^2 n}, \frac{n}{k^2\log^2n}) \bigr)$, which is an exponential improvement. This analysis is given in Appendix \ref{orthonormal_case}.

\begin{Theorem}[Provably correct descent, informal]
  \label{result_thm_neural_algorithm}
  There exists a neurally plausible algorithm for double-sparse
  coding that converges to $A^*$ with
  geometric rate 
  when the initial estimate $A^0$ has the correct support
  and $(\delta, 2)$-near to $A^*$. The running time per
  iteration is $O(mkp+mrp)$ and the sample
  complexity is $\Otilde(m + \sigmae^2\frac{mnr}{k})$.
\end{Theorem}

Similar to~\citet{arora15-neural}, our proposed algorithm enjoys neural plausibility. Moreover, we can achieve a better running time and sample complexity per iteration than previous methods, particularly in the noisy case. We show in
Appendix~\ref{appdx_improve_arora} that in this regime the sample complexity of~\citet{arora15-neural} is
$\Otilde(m + \sigmae^2\frac{mn^2}{k})$. For instance, when $\sigma_\varepsilon\asymp n^{-1/2}$, the sample complexity bound is significantly worse than $\Otilde(m)$ in the noiseless case. In contrast, our proposed method leverages the sparse structure to overcome this problem and obtain improved results.

We are now ready to introduce our methods in detail. As discussed above, our
approach consists of two stages: an initialization algorithm that
produces a coarse estimate of $A^*$, and a descent-style algorithm
that refines this estimate to accurately recover $A^*$.


%% file: initialization.tex
\section{Stage 1: Initialization}
\label{init}
In this section, we present a neurally plausible algorithm that can produce
a coarse initial estimate
of the ground truth $A^*$. We give a neural implementation of the algorithm in Appendix~\ref{neural_implementation}.
 
Our algorithm is an adaptation from the algorithm in~\citet{arora15-neural}. The idea is to estimate dictionary atoms in a greedy fashion by
iteratively re-weighting the given samples. The samples are re-scaled in a
way that the weighted (sample) covariance matrix has the dominant first
singular value, and its corresponding eigenvector is close to
one particular atom with high probability.  However, while this algorithm is conceptually very appealing, it incurs severe computational costs in practice. More precisely, the overall running time is $\Otilde(mn^2p)$ in expectation, which is unrealistic for large-scale problems.  

To overcome this burden, we leverage the double-sparsity assumption in our generative model to obtain a more efficient approach. The high-level idea is to first estimate the support of each column in the synthesis matrix $A^*$, and then obtain a coarse estimate of the nonzero coefficients of each column based on knowledge of its support. The key ingredient of our method is a novel spectral procedure that gives us an estimate of the column supports purely from the observed samples. The full algorithm, that we call \emph{Truncated Pairwise Reweighting}, is listed in pseudocode form below as Algorithm \ref{alg_neural_initialization}.

\begin{algorithm}[!t] 
  \begin{algorithmic} 
    \State \textbf{Initialize} $L = \emptyset$
    \State Randomly divide $p$ samples into two disjoint sets $\mathcal{P}_1$ and $\mathcal{P}_2$ of sizes $p_1$ and $p_2$ respectively
    \State \textbf{While} $|L| < m$. Pick $u$ and $v$ from $\mathcal{P}_1$ at random
    \State \indent For every $l = 1, 2, \dots, n$; compute $$\ehat_{l} =
    \frac{1}{p_2}\sum_{i=1}^{p_2}\inprod{y^{(i)}}{u}
    \inprod{y^{(i)}}{v}(y_l^{(i)})^2$$
    \State \indent Sort $(\ehat_1, \ehat_2, \dots, \ehat_n)$ in descending order
    \State \indent \textbf{If} $r' \leq r$ s.t $\ehat_{(r')} \geq O(k/mr)$ and $\ehat_{(r'+1)}/\ehat_{(r')} < O^*(r/\log^2n)$
    \State \indent \indent Let $\Rhat$ be set of the $r$ largest entries of $\ehat$
    \State \indent \indent $\Mhuv = \frac{1}{p_2}\sum_{i=1}^{p_2}\langle y^{(i)}, u\rangle \langle y^{(i)}, v\rangle y^{(i)}_{\Rhat}(y^{(i)}_{\Rhat})^T$
    \State \indent \indent $\delta_1, \delta_2 \leftarrow$ top singular values of $\Mhuv$
    \State \indent \indent  $z_{\Rhat} \leftarrow$ top singular vector of $\Mhuv$ 
    \medskip
    \State \indent \indent \textbf{If} $\delta_1 \geq \Omega(k/m)$ and $\delta_2 < O^*(k/m\log n)$
    \medskip
    \State \indent \indent \indent \textbf{If} $\text{dist}(\pm z,l) > 1/\log n $ for any $l \in L$ 
    \medskip
    \State \indent \indent \indent \indent Update $L = L \cup \{z\}$
    \State \textbf{Return} $A^0 = (L_1, \dots, L_m)$ 
  \end{algorithmic}
\caption{Truncated Pairwise Reweighting}
\label{alg_neural_initialization}
\end{algorithm}

Let us provide some intuition of our algorithm. Fix a sample $y = A^* x^* + \varepsilon$ from the available training set, and consider samples 
$$u = A^*\alpha + \varepsilon_u, v = A^*\alpha' + \varepsilon_v.$$ 
Now, consider the (very coarse) estimate for the sparse code of $u$ with respect to $A^*$:  
$$\beta = A^{*T}u = A^{*T}A^*\alpha + A^{*T}\varepsilon_u.$$ 
As long as $A^*$ is incoherent enough and $\varepsilon_u$ is small, the estimate $\beta$ behaves just like
$\alpha$, in the sense that for each sample $y$: 
$$\inprod{y}{u} \approx \inprod{x^*}{\beta} \approx \inprod{x^*}{\alpha}.$$ 
Moreover, the above inner products are large only if $\alpha$ and $x^*$ share some elements in their supports; else, they are likely to be small. Likewise, the weight $\inprod{y}{u}\inprod{y}{v}$ depends on whether or not $x^*$ shares the support with both $\alpha$ and $\alpha'$. 

Now, suppose that we have a mechanism to isolate pairs $u$ and $v$ who share
exactly one atom among their sparse representations. Then by scaling each sample $y$ with an increasing function of
$\inprod{y}{u}\inprod{y}{v}$ and linearly adding the samples, we magnify the importance of the samples that are aligned with that atom, and diminish the rest. The final direction can be obtained via the top \emph{principal component} of the reweighted samples and hence can be used as a coarse estimate of the atom. This is exactly the approach adopted in~\citet{arora15-neural}. However, in our double-sparse coding setting, we know that the estimated atom should be sparse as well. Therefore, 
we can naturally perform an extra ``sparsification'' step of the output. An extended algorithm and its correctness are provided in Appendix~\ref{appdx_improve_arora}. However, as we discussed above, the computational complexity of the re-weighting step still remains. 

We overcome this obstacle by first identifying the locations of the nonzero
entries in each atom. Specifically, define the matrix:
$$M_{u,v} = \frac{1}{p_2}\sum_{i=1}^{p_2}\inprod{y^{(i)}}{u}
\inprod{y^{(i)}}{v}y^{(i)}\circ y^{(i)}.$$ 
Then, the diagonal entries of $M_{u,v}$ reveals
the support of the atom of $A^*$ shared among $u$ and $v$: the $r$-largest entries of $M_{u,v}$ will correspond to the support we seek. 
Since the desired direction remains unchanged in the 
$r$-dimensional subspace of its nonzero elements, we can restrict our attention to this subspace, construct a reduced covariance matrix $\Mhuv$, and proceed as before. 
This truncation step alleviates the computational burden by a significant amount; the running time is now $\Otilde(mnp)$, which improves the
original by a factor of $n$.

The success of the above procedure relies upon whether or not we can isolate pairs $u$ and $v$
that share one dictionary atom. Fortunately, this can be done via
checking the decay of the singular values of the (reduced) covariance
matrix. Here too, we show via our analysis that the truncation step plays an important role. Overall, our proposed algorithm not only accelerates the initialization in terms of running time, but also improves the sample complexity over~\cite{arora15-neural}. The performance of Algorithm~\ref{alg_neural_initialization} is described in the following theorem, whose formal proof is deferred to Appendix~\ref{appdx_initialization_analysis}.
 
\begin{Theorem}
\label{main_thm_initialization}
  Suppose that Assumptions \textnormal{\textbf{B1-B4}} hold and
  Assumptions \textnormal{\textbf{A1-A3}} satify with $\mu =
  O^*\bigl(\frac{\sqrt{n}}{k\log^3n}\bigr)$
  and $r = O^*(\log n)$. When $p_1 = \Omgtilde(m)$ and $p_2 =
  \Omgtilde(mr)$, then with high probability Algorithm \ref{alg_neural_initialization} returns an initial estimate $A^0$ whose columns share the same support as $A^*$ and with $(\delta, 2)$-nearness to $A^*$ with $\delta = O^*(1/\log n)$.
\end{Theorem}
%

The limit on $r$ arises from the minimum non-zero coefficient $\tau$ of $A^*$.
Since the columns of $A^*$ are standardized, $\tau$ should degenerate as $r$ grows. In other words, it is getting harder to distinguish the ``signal'' coefficients from zero as $r$ grows with $n$.
However, this limitation can be relaxed when a better incoherence available, for example the orthonormal case. We study this in Appendix~\ref{orthonormal_case}.

To provide some intuition about the working of the algorithm (and its proof), let us analyze it in the case where we have access to infinite number of samples. This setting, of course, is unrealistic. However, the analysis is much simpler and more transparent since we can focus on expected values rather than empirical averages. Moreover, the analysis reveals several key lemmas, which we will reuse extensively for proving Theorem \ref{main_thm_initialization}. First, we give some intuition behind the definition of the ``scores'', $\ehat_l$.

\begin{Lemma} 
Fix samples $u$ and $v$ and suppose that $y = A^*x^* + \varepsilon$ is a random sample independent of $u,v$. 
The expected value of the score for the $l^{\textrm{th}}$ component of $y$ is given by:
  \begin{align*}
    e_l \triangleq  \E[\inprod{y}{u}\inprod{y}{v} y_l^2] 
    = \sum_{i \in U \cap V}q_ic_i\beta_i\beta'_iA^{*2}_{li} + ~\text{perturbation terms}
  \end{align*}
where $q_i = \Prob[i \in S]$, $q_{ij} = \Prob[i, j \in S]$ and $c_i =
\E[x_i^4|i \in S]$. Moreover, the perturbation terms have absolute value at most $O^*(k/m \log n)$.
\label{lm_diagonal_entries_in_expectation}
\end{Lemma}

From \AsmpB{1}, we know that $q_i= \Theta(k/m)$, $q_{ij} =
\Theta(k^2/m^2)$ and $c_i = \Theta(1)$. Besides, we will show later
that $\abs{\beta_i} \approx \abs{\alpha_i} = \Omega(1)$ for $i \in U$, and $\abs{\beta_i} = o(1)$ for $i \notin
U$. Consider the first term $E_0 = \sum_{i \in U \cap V}q_ic_i\beta_i\beta'_iA^{*2}_{li}$. 
Clearly, $E_0 = 0$ if $U \cap V = \emptyset$ or that $l$ does not belong to support of any atom in $U \cap V$. On the contrary, as $E_0 \neq 0$ and $U\cap V = \{i\}$ , then $E_0 = \abs{q_ic_i\beta_i\beta'_iA^{*2}_{li}} \geq
\Omega(\tau^2k/m) = \Omega(k/mr)$ since $\abs{q_ic_i\beta_i\beta'_i} \geq
\Omega(k/m)$ and $\abs{A_{li}^*} \geq \tau$.

Therefore, Lemma \ref{lm_diagonal_entries_in_expectation} suggests that if
$u$ and $v$ share a unique atom among their sparse representations, and $r$ is not too large, then we can indeed recover the correct support of the shared atom. When this is the case, the expected scores corresponding to the nonzero elements of the shared atom will dominate the remaining of the scores. 

Now, given that we can isolate the support $R$ of the corresponding atom, the remaining questions are how best we can estimate its non-zero coefficients, and when $u$ and $v$ share a unique elements in their supports. These issues are handled in the following lemmas.

\begin{Lemma}
  \label{lm_reweighted_cov_matrix_in_expectation}
  Suppose that $u = A^*\alpha + \varepsilon_u$ and $v = A^*\alpha' + \varepsilon_v$ are two random
  samples. Let $U$ and $V$ denote the supports of $\alpha$ and
  $\alpha'$ respectively. $R$ is the support of some atom of interest. The truncated re-weighting matrix is formulated as
\begin{align*}
  \Muv &\triangleq \E[\inprod{y}{u}\inprod{y}{v} y_Ry_R^T]
  = \sum_{i \in U \cap V}q_ic_i\beta_i\beta'_i\AR{i}^*\AR{i}^{*T} + ~\text{perturbation terms}
\end{align*}
where the perturbation terms have norms at most $O^*(k/m\log n)$.
\end{Lemma}

Using the same argument for bounding $E_0$ in Lemma \ref{lm_diagonal_entries_in_expectation}, we can see that $M_0 \triangleq q_ic_i\beta_i\beta'_i\AR{i}^*\AR{i}^{*T}$ has norm at least $\Omega(k/m)$ when $u$ and $v$ share a unique element $i$ ($\norm{\AR{i}^*} = 1$). According to this lemma, the spectral norm of $M_0$ dominates those of the other perturbation terms. Thus, given $R$ we can use the first singular vector of $\Muv$ as an estimate of $\cAi^*$.

\begin{Lemma}
  \label{lm_support_consists_and_closeness}
  Under the setup of Theorem \ref{main_thm_initialization},
  suppose $u = A^*\alpha + \varepsilon_u$ and $v = A^*\alpha' +
  \varepsilon_v$ are two random samples with supports $U$ and $V$
  respectively. $R =  \supp(A^*_i)$. If $u$ and $v$ share the unique
  atom $i$, the first $r$ largest entries of $e_l$ is at least
  $O(k/mr)$ and belong to $R$. Moreover, the top singular vector of $\Muv$ is $\delta$-close to $\AR{i}^*$ for $O^*(1/\log n)$.
\end{Lemma}

\proof The recovery of $\cAi^*$'s support directly follows Lemma \ref{lm_diagonal_entries_in_expectation}. For the latter part, recall from Lemma \ref{lm_reweighted_cov_matrix_in_expectation} that
$$\Muv = q_ic_i\beta_i\beta'_i\AR{i}^*\AR{i}^{*T} + ~\text{perturbation terms}$$
The perturbation terms have norms bounded by $O^*(k/m\log n)$. On the other hand, the first term is has norm at least $\Omega(k/m)$ since $\norm{\AR{i}^*} = 1$ for the correct support $R$ and $\abs{q_ic_i\beta_i\beta'_i} \geq \Omega(k/m)$. Then using Wedin's Theorem to $M_{u,v}$, we can conclude that the top singular vector must be $O^*(k/m\log n)/\Omega(k/m) = O^*(1/\log n)$ -close to $\AR{i}^*$. \qed

\begin{Lemma}
  \label{lm_condition_uv_share_unique_supp}
  Under the setup of Theorem \ref{main_thm_initialization},
  suppose $u = A^*\alpha + \varepsilon_u$ and $v = A^*\alpha' +
  \varepsilon_v$ are two random samples with supports $U$ and $V$ respectively. If the top singular value of $M_{u,v}$ is at least $\Omega(k/m)$ and the second largest one is at most $O^*(k/m\log n)$, then $u$ and $v$ share a unique dictionary element with high probability. 
\end{Lemma}
\proof The proof follows from that of Lemma 37 in~\citet{arora15-neural}. The main idea is to separate the possible cases of how $u$ and $v$ share support and to use Lemma \ref{lm_reweighted_cov_matrix_in_expectation} with the bounded perturbation terms to conclude when $u$ and $v$ share exactly one. We note that due to the condition where $\ehat_{(s)} \geq \Omega(k/mr)$ and $\ehat_{(s+1)}/\ehat_{(s)} \leq O^*(r/\log n)$, it must be the case that $u$ and $v$ share only one atom or share more than one atoms with the same support. When their supports overlap more than one, then the first singular value cannot dominate the second one, and hence it must not be the case. \qed

Similar to~\citep{arora15-neural}, our initialization algorithm requires $\Otilde(m)$ iterations in expectation to estimate all the atoms, hence the expected running time is $\Otilde(mnp)$. All the proofs of Lemma \ref{lm_diagonal_entries_in_expectation} and \ref{lm_reweighted_cov_matrix_in_expectation} are deferred to Appendix \ref{appdx_initialization_analysis}.



%% file: algorithm.tex
\section{Stage 2: Descent}
\label{descent}

We now adapt the neural sparse coding approach of \citet{arora15-neural} to obtain an improved estimate of $A^*$. As mentioned earlier, at a high level the algorithm is akin to performing approximate gradient descent. The insight is that within a small enough neighborhood (in the sense of $\delta$-closeness) of the true $A^*$, an estimate of the ground-truth code vectors, $X^*$, can be constructed using a neurally plausible algorithm. 

The innovation, in our case, is the double-sparsity model since we know \emph{a priori} that $A^*$ is itself sparse. Under sufficiently many samples, the support of $A^*$ can be deduced from the initialization stage; therefore we perform an extra \emph{projection} step in each iteration of gradient descent. In this sense, our method is non-trivially different from \cite{arora15-neural}. The full algorithm is presented as Algorithm \ref{alg_neural_doubly_sdl}.

As discussed in Section \ref{setup}, convergence of noisy approximate gradient descent can be achieved as long as $\ghat^s$ is correlated-w.h.p. with the true solution. However, an analogous convergence result for \emph{projected} gradient descent does not exist in the literature. We fill this gap via a careful analysis.
Due to the projection, we only require the correlated-w.h.p.~property for \emph{part} of $\ghat^s$ (i.e.,~when it is restricted to some support set)
with $A^*$. The descent property is still achieved via Theorem \ref{main_thm_columnwise_descent_in_expectation}. 
Due to various perturbation terms, $\ghat$ is only a biased estimate of $\nabla_A\mathcal{L}(A, X)$; therefore, we can only refine the estimate of $A^*$ until the column-wise error is of order $O(\sqrt{k/n})$. The performance of Algorithm~\ref{alg_neural_doubly_sdl} can be characterized via the following theorem.

\begin{algorithm}[t] 
  \begin{algorithmic} 
    \State \textbf{Initialize} $A^0$ is $(\delta, 2)$-near to $A^*$. $H = (h_{ij})_{n\times m}$ where $h_{ij} = 1$ if $i \in
    \supp(\cA{j}^0)$ and 0 otherwise.
    \State \textbf{Repeat} for $s = 0, 1, \dots, T$ 
    \State \indent Encode: $x^{(i)} = \thres_{C/2}((A^{s})^Ty^{(i)})$ \quad for $i = 1, 2, \dots, p$
    \State \indent Update: $A^{s+1} =  \mathcal{P}_{H}(A^s - \eta\ghat^s) = A^s - \eta\mathcal{P}_{H}(\ghat^s)$
    \State \indent  where $\ghat^s = \frac{1}{p}\sum_{i=1}^p (A^{s}x^{(i)} - y^{(i)})\sgn(x^{(i)})^T$  and $\mathcal{P}_{H}(G) = H \circ G$ 
  \end{algorithmic}
\caption{Double-Sparse Coding Descent Algorithm}
\label{alg_neural_doubly_sdl}
\end{algorithm}
\begin{Theorem}
  \label{main_thm_columnwise_descent_in_expectation}
  Suppose that the initial estimate $A^0$ has the correct column
  supports and is $(\delta, 2)$-near to $A^*$ with $\delta = O^*(1/\log n)$. If Algorithm \ref{alg_neural_doubly_sdl} is provided with $p = \Omgtilde(mr)$ fresh samples at each step and $\eta = \Theta(m/k)$, then
  $$\E[\norm{\cAi^s - \cAi^*}^2] \leq (1-\rho)^s \norm{\cAi^0 - \cAi^*}^2 + O(\sqrt{k/n}) $$
  for some $0 < \rho < 1/2$ and for $s = 1, 2, \dots,
  T$. Consequently, $A^s$ converges to $A^*$ geometrically until column-wise error $O(\sqrt{k/n})$.
\end{Theorem}
We defer the full proof of Theorem
\ref{main_thm_columnwise_descent_in_expectation} to Section \ref{sample_complexity}.
In this section, we take a step towards
understanding the algorithm by analyzing $\ghat^s$ in the infinite
sample case, which is equivalent to its expectation $g^s
\triangleq \E[(A^sx - y)\sgn(x)^T]$. We establish the $(\alpha, \beta,
\gamma_s)$-correlation of a truncated version of $\cgi^s$ with
$\cAi^*$ to obtain the descent in Theorem
\ref{thm_columnwise_descent_infinite_samples} for the infinite sample
case.
\begin{Theorem}
  \label{thm_columnwise_descent_infinite_samples}
  Suppose that the initial estimate $A^0$ has the correct column
  supports and is $(\delta, 2)$-near to $A^*$. If Algorithm \ref{alg_neural_doubly_sdl} is provided with infinite number of samples at each step and $\eta = \Theta(m/k)$, then
  $$\norm{\cAi^{s+1} - \cAi^*}^2 \leq (1-\rho) \norm{\cAi^s - \cAi^*}^2 + O\bigl(k^2/n^2 \bigr) $$
  for some $0 < \rho < 1/2$ and for $s = 1, 2, \dots,
  T$. Consequently, it converges to $A^*$ geometrically until column-wise error is $O(k/n)$.
\end{Theorem}

Note that the better error $O(k^2/n^2)$ is due to the fact that infinitely many
samples are given. The term $O(\sqrt{k/n})$ in Theorem
\ref{main_thm_columnwise_descent_in_expectation} is a trade-off between the accuracy and the sample complexity of the algorithm. The proof of this theorem composes of two steps with two  main results: 1) an explicit form of $g^s$ (Lemma \ref{thm_columnwise_descent_infinite_samples}); 2) $(\alpha, \beta, \gamma_s)$-correlation of
column-wise $g^s$ with $A^*$ (Lemma \ref{lm_correlation_gs}). The proof of those lemmas are deferred to Appendix \ref{appdx_main_algorithm}. Since the correlation primarily relies on the $(\delta, 2)$-nearness of $A^s$ to $A^*$ that is provided initially and maintained at each step, then we
need to argue that the nearness is preserved after each step.

\begin{Lemma}
  \label{lm_expected_columwise_update}
  Suppose that the initial estimate $A^0$ has the correct column
  supports and is $(\delta, 2)$-near to $A^*$. The column-wise update has the form $g_{R, i}^s = p_iq_i(\lambda^s_i\AR{i}^s -
\AR{i}^* + \xi_i^s \pm \zeta)$ where $R = \supp(\cAi^s)$, $\lambda_i^s = \inprod{\cAi^s}{\cAi^*}$ and 
  $$\xi_i^s = \AR{-i}^s\diag(q_{ij})(\cA{-i}^s)^T\cAi^*/q_i.$$
Moreover, $\xi_i$ has norm bounded by $O(k/n)$ for $\delta = O^*(1/\log n)$ and $\zeta$ is negligible.
\end{Lemma}
We underline that the correct support of $A^s$ allows us to obtain the
closed-form expression of $g_{R_i, i}^s$ in terms of $\cAi^s$ and
$\cAi^*$. Likewise, the expression (\ref{eq_expected_columwise_update}) suggests that
$\cgi^s$ is almost equal to $p_iq_i(\cAi^s - \cAi^*)$ (since
$\lambda_i^s \approx 1$), which directs to the desired solution $\cAi^*$. 
With Lemma \ref{lm_expected_columwise_update}, we will prove the
$(\alpha, \beta, \gamma_s)$-correlation of the approximate gradient to
each column $\cAi^*$ and the nearness of each new update to the true
solution $A^*$.

\subsection{$(\alpha, \beta, \gamma_s)$-Correlation}
\begin{Lemma}
  \label{lm_correlation_gs}
  Suppose that $A^s$ to be $(\delta, 2)$-near to $A^*$ and $R = \supp(\cAi^*)$,
  then $2g_{R, i}^s$ is $(\alpha, 1/2\alpha,
  \epsilon^2/\alpha)$-correlated with $\AR{i}^*$; that is
  \begin{equation*}
    \inprod{2g_{R, i}^s} {\AR{i}^s - \AR{i}^*} \geq \alpha \Vert \AR{i}^s - \AR{i}^*\Vert^2 + {1}/{(2\alpha)} \Vert g_{R, i}^s\Vert^2 - {\epsilon^2}/{\alpha} 
  \end{equation*}
Futhermore, the descent is achieved by
  $$\norm{\cAi^{s+1} - \cAi^*}^2 \leq (1-2\alpha\eta)^s\norm{\cAi^0 -
    \cAi^*}^2 + \eta\epsilon_s^2/\alpha$$
 where $\delta = O^*(1/\log n)$ and $\epsilon = O\bigl(
   \frac{k^2}{mn} \bigr)$.
\end{Lemma} 

\proof Throughout the proof, we omit the superscript $s$ for simplicity and denote $2\alpha = p_iq_i$. First, we rewrite $\cgi^s$ as a combination of the true direction $\cAi^s - \cAi^*$ and a term with small norm:
\begin{equation}
  \label{eq:3}
  g_{R, i} 
  = 2\alpha(\AR{i} - \AR{i}^*) + v,
\end{equation}
where $v = 2\alpha[(\lambda_i - 1)\cAi + \epsilon_i]$ with norm
bounded. In fact, since $\cAi$ is $\delta$-close to $\cAi^*$, and both
have unit norm, then $\norm{2\alpha(\lambda_i - 1)\cAi}  = \alpha
\norm{ \cAi - \cAi^*}^2 \leq \alpha \norm{\cAi - \cAi^*}$ and $\norm
{\xi_i} \leq O(k/n)$ from the inequality (\ref{eqn:eps_norm}). Therefore,
\begin{equation*}
  \norm{v}  = \norm{2\alpha(\lambda_i - 1)\AR{i} + 2\alpha \xi_i} \leq \alpha \norm{\AR{i} - \AR{i}^*} + \epsilon
\end{equation*}
where $\epsilon = O(k^2/mn)$. Now, we make use of (\ref{eq:3}) to show the first part of Lemma \ref{lm_correlation_gs}:
\begin{align}
  \inprod{2g_{R, i}}{\AR{i} - \AR{i}^*} = 4\alpha \norm{\AR{i} - \AR{i}^*}^2 + \inprod{2v}{\AR{i} - \AR{i}^*}.
  \label{eq:4}
\end{align}
We want to lower bound the inner product term with respect to $\norm{g_{R_i, i}}^2$ and $\norm{\AR{i} - \AR{i}^*}^2$. Effectively, from \eqref{eq:3}
\begin{align}
  4\alpha \inprod{v}{ \cAi - \cAi^*} &= \norm{g_{R, i}}^2 - 4\alpha^2\norm{\AR{i} - \AR{i}^*}^2 - \norm{v}^2 \nonumber \\
                                      &\geq \norm{g_{R, i}}^2 - 6\alpha^2\norm{\AR{i} - \AR{i}^*}^2 - 2\epsilon^2,
\label{eq:5}
\end{align}
where the last step is due to Cauchy-Schwarz inequality: $\norm{v}^2 \leq 2(\alpha^2 \norm{\AR{i} - \AR{i}^*}^2 + \epsilon^2)$.

Substitute $2\inprod{v}{ \cAi - \cAi^*}$ in \eqref{eq:4} for the right hand side of \eqref{eq:5}, we get the first result: 
$$ \inprod{2g_{R, i}}{\AR{i} - \AR{i}^*} \geq \alpha \norm{\AR{i} - \AR{i}^*}^2
+ \frac{1}{2\alpha}\norm{g_{R, i}}^2 - \frac{\epsilon^2}{\alpha}.$$
The second part is directly followed from Theorem
\ref{thm_descent_from_correlation_z}. Moreover, we have $p_i =
\Theta(k/m)$ and $q_i = \Theta(1)$, then $\alpha = \Theta(k/m)$,
$\beta = \Theta(m/k)$ and $\gamma_s = O(k^3/mn^2)$. Then $g_{R, i}^s$
is $(\Omega(k/m), \Omega(m/k), O(k^3/mn^2))$-correlated with the true
solution $\AR{i}^*$. \qedhere

\proof[Proof of Theorem \ref{thm_columnwise_descent_infinite_samples}] The descent in Theorem \ref{thm_columnwise_descent_infinite_samples} directly follows from the above lemma. Next, we will establish the nearness for the update at step $s$:

\subsection{Nearness}
\begin{Lemma}
  \label{lm_nearness_infinite_sample}
  Suppose that $A^s$ is $(\delta, 2)$-near to $A^*$, then $\norm{A^{s+1} - A^*} \leq 2\norm{A^*} $
\end{Lemma}

\proof[Proof] From Lemma \ref{lm_expected_columwise_update} we have $\cgi^s = p_iq_i(\lambda_i\cAi^s - \cAi^*) + \cB{-i}\diag(q_{ij})\cB{-i}^T\cAi^* \pm \zeta$. Denote $\bar{R} = [n] \backslash R$, then it is obvious that $g_{\bar{R}, i}^s = A_{\bar{R}, -i}\diag(q_{ij})\cB{-i}^T\cAi^* \pm \zeta$ is bounded by $O(k^2/m^2)$. Then we follows the proof of Lemma 24 in~\citep{arora15-neural} for the nearness with full $g^s = g_{R,i}^s + g_{\bar{R}, i}^s$ to finish the proof for this lemma. \qed


In sum, we have shown the descent property of Algorithm~\ref{alg_neural_doubly_sdl} in the infinite sample case. The study of the concentration of $\ghat^s$ around its mean to the sample complexity is provided in Section~\ref{sample_complexity}. In the next section, we corroborate our theory by some numerical results on synthetic data.


%% file: empirical.tex
\input{plots_tikz}

\section{Empirical Study}
\label{empirical}


We compare our method with three different methods for both standard sparse and
double-sparse coding. For the standard approach, we implement the
algorithm proposed in~\citet{arora15-neural}, which currently is the best theoretically sound method for provable sparse coding.
However, since their method does not explicitly leverage the double-sparsity model, we also implement a heuristic modification that performs a hard thresholding (HT)-based post-processing step in the initialization and learning procedures (which we dub \emph{Arora + HT}). The
final comparison is the \emph{Trainlets} approach of~\citet{sulam16-trainlets}. 

We generate a synthetic training dataset according to the model described in Section \ref{setup}. The base dictionary $\Phi$ is the identity matrix of
size $n =64$ and the square synthesis matrix $A^*$ is a block diagonal matrix with 32 blocks. Each $2 \times 2$ block is of form $[1~1; 1~-1]$ 
(i.e., the column sparsity $r = 2$) 
. The support of $x^*$ is drawn
uniformly over all $6$-dimensional subsets of $[m]$,
 and the nonzero coefficients are randomly set to $\pm 1$ with equal probability.
In our simulations with noise, we add Gaussian noise
$\varepsilon$ with entrywise variance $\sigmae^2 = 0.01$ 
to each of
those above samples. 
For all the approaches except {\trainlets}, we use $T = 2000$ iterations for the initialization procedure, and set the number of steps in the descent stage to 25. Since {\trainlets} does not have a specified initialization procedure, we initialize it with a random Gaussian matrix upon which column-wise sparse thresholding is then performed. The learning step of {\trainlets}\footnote{We utilize {\trainlets}'s implementation provided at {http://jsulam.cswp.cs.technion.ac.il/home/software/.}} is executed for $50$ iterations, which tolerates its initialization deficiency. For each Monte Carlo trial, we uniformly draw $p$ samples, feed these samples to the four different algorithms, and observe their ability to reconstruct $A^*$. Matlab implementation of our algorithms is available online\footnote{https://github.com/thanh-isu/double-sparse-coding}.

We evaluate these approaches on three metrics as a function of the number of available samples: (i) fraction of trials in which each algorithm successfully recovers the ground truth $A^*$; (ii) reconstruction error; and (iii) running time. The synthesis matrix is said to be ``successfully recovered'' if the Frobenius norm of the difference between the estimate $\widehat{A}$ and the ground truth $A^*$ is smaller than a threshold
which is set to $10^{-4}$ in the noiseless case, and to $0.5$ in the other. 
 All three metrics are averaged over 100 Monte Carlo simulations.
As discussed above, the Frobenius norm is only meaningful under a suitable permutation and sign flip transformation linking $\widehat{A}$ and $A^*$. We estimate this transformation using a simple maximum weight matching algorithm. Specifically, we construct a weighted bipartite graph with nodes representing columns of $A^*$ and $\widehat{A}$ and adjacency matrix defined as $G = \abs{A^{*T}\widehat{A}}$, where $\abs{\cdot}$ is taken element-wise. We compute the optimal matching using the Hungarian algorithm, and then estimate the sign flips by looking at the sign of the inner products between the matched columns.

The results of our experiments are shown in Figure \ref{fig_simulation_results} with the
top and bottom rows respectively for the noiseless and noisy cases. 
The two leftmost figures suggest that all algorithms exhibit a ``phase transitions'' in sample complexity that
occurs in the range of 500-2000 samples. 
In the noiseless case, our method achieves the phase transition with the fewest number of samples. 
In the noisy case, our method nearly matches the best sample complexity performance (next to \trainlets, which is a heuristic and computationally expensive).
Our method achieves the best performance in terms of (wall-clock) running time in all cases.


%% file: plots_tikz.tex
\begin{figure*}[!t]
\centering
\begin{tabular}{ccc}

\begin{tikzpicture}[scale=0.9]
\begin{axis}[
		width=3.75cm,
		height=3cm,
		scale only axis,
		xmin=0, xmax=5000,
		xlabel = {Sample size},
		xmajorgrids,
		ymin=0, ymax=1,
		ylabel={Recovery rate},
		ymajorgrids,
		line width=1.0pt,
		mark size=1.5pt,
		legend style={nodes={scale=0.75, transform shape},at={(0.50,.01)},anchor=south west,draw=black,fill=white,align=left}
		]
\addplot  [color=red,
		solid, 
		very thick,
		mark=o,
		mark options={solid,scale=1.5},
		]
		table [x index = 0,y index=1]{results/prob_success_nless.txt};
\addplot [color=black,
		solid, 
		very thick,
		mark=square,
		mark options={solid,scale=1.5},
		]
		table [x index = 0,y index=2]{results/prob_success_nless.txt};
\addplot [color=orange,
		solid, 
		very thick,
		mark= diamond*,
		mark options={solid,scale=1.5},
		]
		table [x index = 0,y index=3]{results/prob_success_nless.txt};
\addplot [color=blue,
		solid, 
		very thick,
		mark=star,
		mark options={solid,scale=1.5},
		]
		table [x index = 0,y index=4]{results/prob_success_nless.txt};

\end{axis}
\end{tikzpicture}
&
\begin{tikzpicture}[scale=0.9]
\begin{axis}[
		width=3.75cm,
		height=3cm,
		scale only axis,
		xmin=0, xmax=5000,
		xlabel = {Sample size},
		xmajorgrids,
		ymin= 0, ymax=8,
		ylabel={Reconstruction error},
		ymajorgrids,
		line width=1.0pt,
		mark size=1.0pt,
		legend style={nodes={scale=0.75, transform shape},at={(0.60,.8)},anchor= north,draw=black,fill=white,align=left}           
		]
\addplot  [color=red,
		solid, 
		very thick,
		mark=o,
		mark options={solid,scale=1.5},
		]
		table [x index = 0,y index=1]{results/error_nless.txt};
\addlegendentry{Ours}
\addplot [color=black,
		solid, 
		very thick,
		mark=square,
		mark options={solid,scale=1.5},
		]
		table [x index = 0,y index=2]{results/error_nless.txt};
\addlegendentry{Arora}
\addplot [color=orange,
		solid, 
		very thick,
		mark= diamond*,
		mark options={solid,scale=1.5},
		]
		table [x index = 0,y index=3]{results/error_nless.txt};
\addlegendentry{Arora+HT}
\addplot [color=blue,
		solid, 
		very thick,
		mark=star,
		mark options={solid,scale=1.5},
		]
		table [x index = 0,y index=4]{results/error_nless.txt};
\addlegendentry{Trainlets}            
\end{axis}
\end{tikzpicture}
&
\begin{tikzpicture}[scale=0.9]
\begin{axis}[
		width=3.75cm,
		height=3cm,
		scale only axis,
		xmin=0, xmax=5000,
		xlabel = {Sample size},
		xmajorgrids,
		ymin= 0, ymax=4,
		ylabel={Running time},
		ymajorgrids,
		line width=1.0pt,
		mark size=1.0pt,
		legend style={nodes={scale=0.75, transform shape},at={(0.40,.01)},anchor=south west,draw=black,fill=white,align=left}           
		]
\addplot  [color=red,
		solid, 
		very thick,
		mark=o,
		mark options={solid,scale=1.5},
		]
		table [x index = 0,y index=1]{results/run_time_nless.txt};
\addplot [color=black,
		solid, 
		very thick,
		mark=square,
		mark options={solid,scale=1.5},
		]
		table [x index = 0,y index=2]{results/run_time_nless.txt};
\addplot [color=orange,
		solid, 
		very thick,
		mark= diamond*,
		mark options={solid,scale=1.5},
		]
		table [x index = 0,y index=3]{results/run_time_nless.txt};
\addplot [color=blue,
		solid, 
		very thick,
		mark=star,
		mark options={solid,scale=1.5},
		]
		table [x index = 0,y index=4]{results/run_time_nless.txt};

\end{axis}
\end{tikzpicture}

\\

\begin{tikzpicture}[scale=0.9]
\begin{axis}[
        width=3.75cm,
        height=3cm,
        scale only axis,
        xmin=0, xmax=5000,
        xlabel = {Sample size},
        xmajorgrids,
        ymin=0, ymax=1,
        ylabel={Recovery rate},
        ymajorgrids,
        line width=1.0pt,
        mark size=1.5pt,
        legend style={nodes={scale=0.75, transform shape},at={(0.50,.01)},anchor=south west,draw=black,fill=white,align=left}
        ]
\addplot  [color=red,
       solid, 
       very thick,
       mark=o,
       mark options={solid,scale=1.5},
       ]
       table [x index = 0,y index=1]{results/prob_success_noise.txt};
\addplot [color=black,
       solid, 
       very thick,
       mark=square,
       mark options={solid,scale=1.5},
       ]
       table [x index = 0,y index=2]{results/prob_success_noise.txt};
\addplot [color=orange,
       solid, 
       very thick,
       mark= diamond*,
       mark options={solid,scale=1.5},
       ]
       table [x index = 0,y index=3]{results/prob_success_noise.txt};
\addplot [color=blue,
       solid, 
       very thick,
       mark=star,
       mark options={solid,scale=1.5},
       ]
       table [x index = 0,y index=4]{results/prob_success_noise.txt};
\end{axis}
\end{tikzpicture}
&
\begin{tikzpicture}[scale=0.9]
\begin{axis}[
        width=3.75cm,
        height=3cm,
        scale only axis,
        xmin=0, xmax=5000,
        xlabel = {Sample size},
        xmajorgrids,
        ymin= 0, ymax=8,
        ylabel={Reconstruction error},
        ymajorgrids,
        line width=1.0pt,
        mark size=1.0pt,
        legend style={nodes={scale=0.75, transform shape},at={(0.60,.8)},anchor= north,draw=black,fill=white,align=left}           
        ]
\addplot  [color=red,
		solid, 
		very thick,
		mark=o,
		mark options={solid,scale=1.5},
		]
		table [x index = 0,y index=1]{results/error_noise.txt};
\addlegendentry{Ours}
\addplot [color=black,
		solid, 
		very thick,
		mark=square,
		mark options={solid,scale=1.5},
		]
		table [x index = 0,y index=2]{results/error_noise.txt};
\addlegendentry{Arora}
\addplot [color=orange,
		solid, 
		very thick,
		mark= diamond*,
		mark options={solid,scale=1.5},
		]
		table [x index = 0,y index=3]{results/error_noise.txt};
\addlegendentry{Arora+HT}
\addplot [color=blue,
		solid, 
		very thick,
		mark=star,
		mark options={solid,scale=1.5},
		]
		table [x index = 0,y index=4]{results/error_noise.txt};
\addlegendentry{Trainlets}            
\end{axis}
\end{tikzpicture}
&
\begin{tikzpicture}[scale=0.9]
\begin{axis}[
		width=3.75cm,
		height=3cm,
		scale only axis,
		xmin=0, xmax=5000,
		xlabel = {Sample size},
		xmajorgrids,
		ymin= 0, ymax=6,
		ylabel={Running time},
		ymajorgrids,
		line width=1.0pt,
		mark size=1.0pt,
		legend style={nodes={scale=0.75, transform shape},at={(0.40,.01)},anchor=south west,draw=black,fill=white,align=left}           
		]
\addplot  [color=red,
		solid, 
		very thick,
		mark=o,
		mark options={solid,scale=1.5},
		]
		table [x index = 0,y index=1]{results/run_time_noise.txt};
\addplot [color=black,
		solid, 
		very thick,
		mark=square,
		mark options={solid,scale=1.5},
		]
		table [x index = 0,y index=2]{results/run_time_noise.txt};
\addplot [color=orange,
		solid, 
		very thick,
		mark= diamond*,
		mark options={solid,scale=1.5},
		]
		table [x index = 0,y index=3]{results/run_time_noise.txt};
\addplot [color=blue,
		solid, 
		very thick,
		mark=star,
		mark options={solid,scale=1.5},
		]
		table [x index = 0,y index=4]{results/run_time_noise.txt};

\end{axis}
\end{tikzpicture}

\end{tabular}
\caption{\small\sl (top) The performance of four methods on three
  metrics (recovery rate, reconstruction error and running time) in sample size in the noiseless case. (bottom) The same metrics are measured for the noisy case. \label{fig_simulation_results}}
\end{figure*}
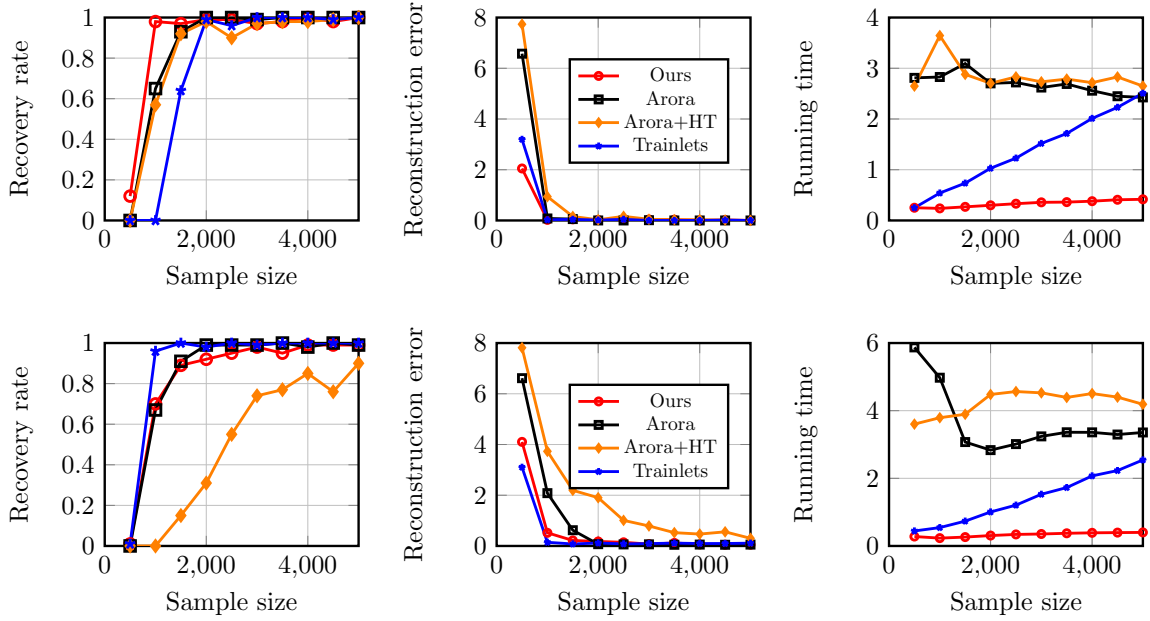

%% file: conclusion.tex
\section{Conclusion}
\label{conc}

In this paper, we have addressed an open theoretical question on learning sparse dictionaries under a special type of generative model.
Our proposed algorithm consists of a novel initialization step followed by a descent-style step,
both are able to take advantage of the sparse structure.
We rigorously demonstrate its efficacy in both sample- and computation-complexity over existing heuristics as well as provable approaches for double-sparse and regular sparse coding. This results in \emph{the first known provable approach for double-sparse coding problem} with statistical and algorithmic guarantees. Besides, we also show three benefits of our approach: neural plausibility, robustness to noise and practical usefulness via the numerical experiments. 

Nevertheless, several fundamental questions regarding our approach remain.
First, our initialization method (in the overcomplete case) achieves its theoretical guarantees under fairly stringent limitations on the sparsity level $r$.
This arises due to our reweighted spectral initialization strategy, and it is an open question whether a better initialization strategy exists (or whether these types of initialization are required at all).
Second, our analysis holds for complete (fixed) bases $\Phi$, and it remains open to study the setting where $\Phi$ is over-complete.
Finally, understanding the reasons behind the very promising practical performance of methods based on heuristics, such as {\trainlets}, on real-world data remains a very challenging open problem.


%% file: appendix.tex
\appendix

\input{appendix_assumptions}
\input{appendix_initialization}
\input{appendix_algorithm}

\input{sample_complexity}
\input{appendix_ortho_dictionary}
\input{appendix_more_arora}
\input{appendix_neural_impl}


%% file: appendix_assumptions.tex
\section{Auxiliary Lemma}
\label{appdx_assumption}



\begin{Claim}[Maximal row $\ell_1$-norm]
  \label{cl_bound_operator_norm_true_A}
  Given that $\norm{A^*}^2_F = m$ and $\norm{A^*} = O(\sqrt{m/n})$, then $\norm{A^{*T}}_{1, 2} = \Theta(\sqrt{m/n})$.
\end{Claim}
\proof Recall the definition of the operator norm:
$$\norm{A^{*T}}_{1, 2} = \sup_{x \neq 0}
\frac{\norm{A^Tx}}{\norm{x}_1} \leq \sup_{x \neq 0}
\frac{\norm{A^Tx}}{\norm{x}} = \norm{A^{*T}} = O(\sqrt{m/n}). $$
Since $\norm{A^*}^2_F = m$, $\norm{A^{*T}}_{1, 2} \geq
\norm{A^*}_F/\sqrt{n} = \sqrt{m/n}$. Combining with the above, we have
$\|A^{*T}\|_{1,2}=\Theta(\sqrt{m/n})$. \qed

Along with Assumptions \textbf{A1} and \textbf{A3}, the above claim implies
the number of nonzero entries in each row is $O(r)$. This Claim is an
important ingredient in our analysis of our initialization algorithm
shown in Section~\ref{init}.


%% file: appendix_initialization.tex
\section{Analysis of Initialization Algorithm}
\label{appdx_initialization_analysis}

\subsection{Proof of Lemma \ref{lm_diagonal_entries_in_expectation}}
\label{prf_lm_diagonal_entries}

The proof of Lemma \ref{lm_diagonal_entries_in_expectation} can be divided into three steps: 1) we first establish useful properties of $\beta$ with respect to $\alpha$; 2) we then explicitly derive $e_l$ in terms of the generative model parameters and $\beta$; and 3) we finally bound the error terms in $E$ based on the first result and appropriate assumptions.

\begin{Claim}
 \label{cl_bound_subgaussian_rv}
 In the generative model, $\norm{x^*} \leq \Otilde(\sqrt{k})$ and $\norm{\varepsilon} \leq \Otilde(\sigmae\sqrt{n})$ with high probability.
\end{Claim}

\proof The claim directly follows from the fact that $x^*$ is a $k$-sparse random vector whose nonzero entries are independent sub-Gaussian with variance $1$. Meanwhile, $\varepsilon$ has $n$ independent Gaussian entries of variance $\sigmae^2$. \qedhere

Despite its simplicity, this claim will be used in many proofs throughout the paper. Note also that in this section we will calculate the expectation over $y$ and often refer probabilistic bounds (\whp) under the randomness of $u$ and $v$.

\begin{Claim} Suppose that $u = A^*\alpha + \varepsilon_u$ is a random sample and $U = \supp(\alpha)$. Let $\beta = A^{*T}u$, then, \whp,
   we have (a) $\abs{\beta_i - \alpha_i} \leq \frac{\mu k\log n}{\sqrt{n}} + \sigmae\log n$ for each $i$  and
  (b) $\norm{\beta} \leq \Otilde(\sqrt{k} + \sigmae\sqrt{n})$.
   \label{cl_bounds_of_beta}
\end{Claim}

\proof The proof mostly follows from Claim 36 of \cite{arora15-neural}, with
an additional consideration of the error $\varepsilon_u$.
Write $W = U \backslash \{i\}$ and observe that
$$\abs{\beta_i - \alpha_i} = \abs{\cAi^{*T}\cA{W}^*\alpha_W + \cAi^{*T}\varepsilon_u} \leq \abs{\inprod{\cA{W}^{*T}\cAi^*}{\alpha_W}} + \abs{\inprod{\cAi^*}{\varepsilon_u}}$$
Since $A^*$ is $\mu$-incoherence, then $\norm{\cAi^{*T}\cA{W}^*} \leq \mu\sqrt{k/n}$. Moreover, $\alpha_W$ has $k-1$ independent sub-Gaussian entries of variance $1$, therefore
$\abs{\inprod{\cA{W}^{*T}\cAi^*}{\alpha_W} } \leq  \frac{\mu k\log
  n}{\sqrt{n}}$ with high probability. Also recall that
$\varepsilon_u$ has independent Gaussian entries of variance
$\sigmae^2$, then $\cAi^{*T}\varepsilon_u$ is Gaussian with the
same variance ($\norm{\cAi^*} = 1$). Hence $\abs{\cAi^{*T}\varepsilon} \leq \sigmae\log
n$ with high probability. Consequently, $\abs{\beta_i - \alpha_i} \leq \frac{\mu k\log n}{\sqrt{n}} + \sigmae\log n$, which is the first part of the claim.

Next, in order to bound $\norm{\beta}$, we express $\beta$ as
$$\norm{\beta} = \norm{A^{*T}\cA{U}^*\alpha_U + A^{*T}\varepsilon_u} \leq
\norm{A^*}\norm{\cA{U}^*}\norm{\alpha_U} +
\norm{A^*}\norm{\varepsilon_u}$$
Using Claim \ref{cl_bound_subgaussian_rv} to get $\norm{\alpha_U} \leq \Otilde(\sqrt{k})$ and $\norm{\varepsilon_u} \leq
\Otilde(\sigmae\sqrt{n})$ \whp, and further noticing that $\norm{\cA{U}^*} \leq \norm{A^*}
\leq O(1)$ 
, we complete the proof for the second part. \qed

Claim \ref{cl_bounds_of_beta} suggests that the
difference between $\beta_i$ and $\alpha_i$ is bounded above by
$O^*(1/\log^2 n)$ \whp\
if $\mu = O^*(\frac{\sqrt{n}}{k\log^3n})$.
Therefore, \whp, $C - o(1) \leq
\abs{\beta_i} \leq \abs{\alpha_i} + o(1) \leq O(\log m)$ for $i \in U$ and $\abs{\beta_i} \leq
O^*(1/\log^2n)$ otherwise. On the other hand, under Assumption \textbf{B4}, $\norm{\beta} \leq \Otilde(\sqrt{k})$ \whp\ We will use these results multiple times in the next few proofs. 

\proof[Proof of Lemma \ref{lm_diagonal_entries_in_expectation}] We decompose $d_l$ into small parts so that the stochastic model $\mathcal{D}$ is made use.
\begin{align*}
  \label{eq_full_decomposition_dl}
  e_l &= \E[\inprod{y}{u}\inprod{y}{v} y_l^2] = \E[\inprod{A^*x^* + \varepsilon}{u} \inprod{A^*x^* + \varepsilon}{v} (\inprod{A^*_{l\cdot}}{x^*} + \varepsilon)^2] \nonumber \\
  &= \E\bigl[\bigl\{\inprod{x^*}{\beta} \inprod{x^*}{\beta'} + x^{*T}(\beta v^T + \beta'u^T)\varepsilon + u^T\varepsilon\varepsilon^Tv\bigr\}\bigl\{\inprod{\rA{l}^*}{x^*}^2 + 2\inprod{\rA{l}^*}{x^*}\varepsilon_l + \varepsilon_l\bigr\}\bigr] \nonumber \\
  &= E_1 + E_2 + \dots + E_9
\end{align*}
where the terms are 
\begin{align}
  \begin{split}
  &E_1 = \E[\inprod{x^*}{\beta} \inprod{x^*}{\beta'} \inprod{\rA{l}^*}{x^*}^2] \\
  &E_2 = 2\E[\inprod{x^*}{\beta} \inprod{x^*}{\beta'} \inprod{\rA{l}^*}{x^*}\varepsilon_l] \\
  &E_3 = \E[\inprod{x^*}{\beta} \inprod{x^*}{\beta'}\varepsilon_l^2] \\
  &E_4 = \E\bigl[\inprod{A^*_{l\cdot}}{x^*}^2 x^{*T}(\beta v^T + \beta'u^T)\varepsilon \bigr] \\
  &E_5 = \E\bigl[\inprod{A^*_{l\cdot}}{x^*} x^{*T}(\beta v^T + \beta'u^T)\varepsilon \varepsilon_l \bigr] \\
  &E_6 = \E\bigl[(\beta v^T + \beta'u^T)\varepsilon \varepsilon_l^2 \bigr] \\
  &E_7 = \E[u^T\varepsilon\varepsilon^Tv\inprod{\rA{l}^*}{x^*}^2] \\
  &E_8 = 2\E[u^T\varepsilon\varepsilon^Tv\inprod{\rA{l}^*}{x^*}\varepsilon_l] \\
  &E_9 = \E[u^T\varepsilon\varepsilon^Tv\varepsilon_l^2]
  \end{split}
\end{align}
Because $x^*$ and $\varepsilon$ are independent and zero-mean, $E_2$
and $E_4$ are clearly zero. Moreover,
$$E_6 = (\beta v^T + \beta'u^T)
\E[\varepsilon \varepsilon_l^2] = 0$$ due to the fact that
$\E[\varepsilon_j\varepsilon_l^2] = 0$, for $j\neq l$, and $\E[\varepsilon_l^3] = 0$. Also, 
$$E_8 = \rA{l}^{*T}\E[x^*]\E\bigl[
u^T\varepsilon\varepsilon^Tv \varepsilon_l\bigr] = 0.$$ We bound the
remaining terms separately in the following claims.

\begin{Claim} In the decomposition \eqref{eq_full_decomposition_dl}, $E_1$ is of the form
  \label{cl_bound_E1}
  \begin{align*}
    E_1  &= \sum_{i \in U \cap V}q_ic_i\beta_i\beta'_iA^{*2}_{li} + \sum_{i \notin U \cap V}q_ic_i\beta_i\beta'_iA^{*2}_{li} + \sum_{j \neq i} q_{ij}(\beta_i\beta'_iA^{*2}_{lj} + 2\beta_i\beta'_jA^*_{li}A^*_{lj})
  \end{align*}
where all those terms except $\sum_{i \in U \cap V}q_ic_i\beta_i\beta'_iA^{*2}_{li}$ have magnitude at most $O^*(k/m\log^2n)$ \whp
\end{Claim}
\proof Using the generative model in Assumptions \textbf{B1}-\textbf{B4}, we have
\begin{align*}
  E_1 &= \E[\inprod{x^*}{\beta} \inprod{x^*}{\beta'} \inprod{\rA{l}^*}{x^*}^2] \\
  &= \E_S\big[\E_{x^*|S}[\sum_{i \in S}\beta_ix_i^*\sum_{i \in S}\beta'_ix_i^*\big(\sum_{i\in S}A^*_{li}x_i^*\big)^2]\big] \\
  &= \sum_{i\in [m]}q_ic_i\beta_i\beta'_iA^{*2}_{li} + \sum_{i,j \in [m], j \neq i} q_{ij}(\beta_i\beta'_iA^{*2}_{lj} + 2\beta_i\beta'_jA^*_{li}A^*_{lj}) \\
  &= \sum_{i \in U \cap V}q_ic_i\beta_i\beta'_iA^{*2}_{li} + \sum_{i \notin U \cap V}q_ic_i\beta_i\beta'_iA^{*2}_{li} + \sum_{j \neq i} q_{ij}(\beta_i\beta'_iA^{*2}_{lj} + 2\beta_i\beta'_jA^*_{li}A^*_{lj}),
\end{align*}
where we have used the $q_i = \Prob[i \in S]$, $q_{ij} = \Prob[i, j
\in S]$ and $c_i = \E[x_i^4|i \in S]$ and Assumptions \textbf{B1}-\textbf{B4}. We now
prove that the last three terms are upper bounded by
$O^*(k/m\log n)$. The key observation is that all these terms typically
involve a quadratic form of the $l$-th row $\rA{l}^*$ whose norm is
bounded by $O(1)$ (by Claim \ref{cl_bound_operator_norm_true_A} and
Assumption \textbf{A4}). Moreover, $\abs{\beta_i\beta_i'}$ is relatively small for $i \notin U \cap V$ while $q_{ij} = \Theta(k^2/m^2)$. For the second term, we apply the Claim \ref{cl_bounds_of_beta} for $i \in [m]\backslash (U \cap V)$ to bound $\abs{\beta_i\beta_i'}$ 
. Assume $\alpha_i = 0$ and $\alpha_i' \neq 0$, then with high probability 
\begin{align*}
  \abs{\beta_i\beta_i'} 
  &\leq \abs{(\beta_i - \alpha_i)(\beta'_i - \alpha'_i)} + \abs{\beta_i\alpha'_i} \leq O^*(1/\log n)
\end{align*}

Using the bound $q_ic_i = \Theta(k/m)$, we have \whp,
\begin{equation*}
  \abs[\Big]{\sum_{i \notin U \cap V}q_ic_i\beta_i\beta'_iA^{*2}_{li}} \leq \max_{i}\abs{q_ic_i\beta_i\beta'_i} \sum_{i \notin U \cap V}A^{*2}_{li} \leq \max_{i}\abs{q_ic_i\beta_i\beta'_i}\norm{A^*}^2_{1,2}  \leq O^*(k/m\log n).
\end{equation*}

For the third term, we make use of the bounds on $\norm{\beta}$ and
$\norm{\beta'}$ from the previous claim where
$\norm{\beta}\norm{\beta'} \leq \Otilde(k)$ \whp, and on $q_{ij} = \Theta(k^2/m^2)$. More precisely, \whp,
\begin{align*}
  \abs[\Big]{\sum_{j \neq i} q_{ij}\beta_i\beta'_iA^{*2}_{lj}} &= \abs[\Big]{\sum_i\beta_i\beta'_i\sum_{j\neq i}q_{ij}A^{*2}_{lj}} \leq \sum_i\abs{\beta_i\beta'_i}\bigl(\sum_{j\neq i}q_{ij}A^{*2}_{lj}\bigr) \\
&\leq (\max_{i\neq j}q_{ij}) \sum_i\abs{\beta_i\beta_i'} \Bigl( \sum_j A^{*2}_{lj} \Bigr)  \leq (\max_{i\neq j}q_{ij}) \norm{\beta}\norm{\beta'}\norm{A^*}^2_{1,2} \leq \Otilde(k^3/m^2),
\end{align*}
where the second last inequality follows from the Cauchy-Schwarz inequality. For the last term, we write it in a matrix form as $\sum_{j \neq i} q_{ij} \beta_i\beta'_jA^*_{li}A^*_{lj} = \rAl^{*T}Q_\beta\rAl^*$ where $(Q_\beta)_{ij} = q_{ij} \beta_i\beta'_j$ for $i \neq j$ and $(Q_\beta)_{ij} = 0$ for $i = j$. Then
\begin{align*}
   \abs{\rAl^{*T}Q_\beta\rAl^*} \leq \norm{Q_\beta}\norm{\rAl^*}^2 \leq  \norm{Q_\beta}_F \norm{A^*}^2_{1,2},
\end{align*}
where $ \norm{Q_\beta}^2_F = \sum_{i\neq j}  q_{ij}^2 \beta_i^2(\beta'_j)^2 \leq (\max_{i\neq j}q_{ij}^2)\sum_i \beta_i^2\sum_j(\beta'_j)^2 \leq (\max_{i\neq j}q_{ij}^2) \norm{\beta}^2 \norm{\beta'}^2$. Ultimately,
$$\abs[\Big]{\sum_{j \neq i} q_{ij} \beta_i\beta'_jA^*_{li}A^*_{lj}} \leq (\max_{i\neq j}q_{ij}) \norm{\beta} \norm{\beta'}\norm{A^*}^2_{1,2}  \leq \Otilde(k^3/m^2).$$
Under Assumption  $k = O^*(\frac{\sqrt{n}}{\log n})$, then $\Otilde(k^3/m^2) \leq O^*(k/m\log^2 n)$. As a result, the two terms above are bounded by the same amount $O^*(k/m\log n)$ \whp, so we complete the proof of the claim. \qed

\begin{Claim}
  \label{cl_bound_Es}
  In the decomposition (\ref{eq_full_decomposition_dl}), $\abs{E_3}$, $\abs{E_5}$, $\abs{E_7}$ and $\abs{E_9}$ is at most $O^*(k/m\log^2n)$.
\end{Claim}
\proof Recall that $\E[x_i^2|S] = 1$ and $q_i = \Prob[i \in S] =
\Theta(k/m)$ for $S = \supp(x^*)$, then
\begin{align*}
  E_3 &= \E[\inprod{x^*}{\beta} \inprod{x^*}{\beta'}\varepsilon_l^2] = \sigmae^2\E_S\bigl[\E_{x^*|S}[\sum_{i, j \in S}\beta_{i}\beta'_{j}x_{i}^*x_{j}^*]\bigr] \\
  &= \sigmae^2\E_S[\sum_{i \in S}\beta_i\beta'_i] = \sum_{i}\sigmae^2q_i\beta_i\beta'_i
\end{align*}
Denote $Q = \diag(q_1, q_2, \dots, q_m)$, then $\abs{E_3} = \abs{\sigmae^2\inprod{Q\beta}{\beta'}} \leq \sigmae^2\norm{Q}\norm{\beta}\norm{\beta'} \leq \Otilde(\sigmae^2 k^2/m) = \Otilde(k^3/mn)$ where we have used $\norm{\beta} \leq \Otilde(\sqrt{k})$ \whp\ and $\sigmae \leq O(1/\sqrt{n})$.
For convenience, we handle the seventh term before $E_5$:
\begin{align*}
  E_7 = \E[u^T\varepsilon\varepsilon^Tv\inprod{\rA{l}^*}{x^*}^2]
  = \E[\inprod{\rA{l}^*}{x^*}^2]u^T\E[\varepsilon\varepsilon^T]v
  = \sum_{i}\sigmae^2\inprod{u}{v}q_iA_{li}^2 = \sigmae^2\inprod{u}{v}\rAl^TQ\rAl
\end{align*}
To bound this term, we use Claim \ref{cl_bound_y} in Appendix \ref{sample_complexity} to have $\norm{u} = \norm{A^*\alpha + \varepsilon_u} \leq \Otilde(\sqrt{k})$ \whp\ and $\inprod{u}{v} \leq \Otilde(\sqrt{k})$ \whp\
Consequently, $\abs{E_7} \leq \sigmae^2
\norm{Q}\norm{\rA{l}}^2\abs{\inprod{u}{v}} \leq \Otilde(k^2/mn)$
because $\norm{\rA{l}}^2 \leq O(m/n)$ and $\sigmae \leq O(1/\sqrt{n})$.
Now, the firth term $E_5$ is expressed as follows
\begin{align*}
  E_5 &= \E\bigl[\inprod{A^*_{l\cdot}}{x^*} x^{*T}(\beta v^T + \beta'u^T)\varepsilon \varepsilon_l \bigr] \\
  &= \rA{l}^{*T}\E\bigl[x^*x^{*T}\bigr] (\beta v^T + \beta'u^T) \E[\varepsilon \varepsilon_l] \\ 
  &= \sigmae^2\rA{l}^{*T}Q(v_l\beta + u_l\beta')
\end{align*}
Observe that $\abs{E_5} \leq \sigmae^2\norm{\rA{l}^{*T}}\norm{Q(v_l\beta + u_l\beta')} \leq \sigmae^2\norm{\rA{l}^{*T}}\norm{Q}\norm{v_l\beta + u_l\beta'}$ and that $\norm{v_l\beta + u_l\beta'} \leq 2\norm{u}\norm{\beta} \leq \Otilde(k)$ \whp\ using the result $\norm{u} \leq \Otilde(k)$ and $\norm{\beta} \leq \Otilde(k)$ from Claim \ref{cl_bounds_of_beta}, then $E_5$ bounded by $\Otilde(k^2/mn)$.

The last term
\begin{align*}
  E_9 &= \E[u^T\varepsilon\varepsilon^Tv\varepsilon_l^2] = u^T\E\bigl[\varepsilon\varepsilon^T \varepsilon_l^2\bigr]v = 9\sigmae^4\inprod{u}{v}
\end{align*}
because the independent entries of $\varepsilon$ and
$\E[\varepsilon_l^4] = 9\sigmae^4$.  Therefore, $\abs{E_9} \leq
9\sigmae^4\norm{u}\norm{v} \leq \Otilde(k^2/n^2)$. Since $m=O(n)$ and $k \leq O^*(\frac{\sqrt{n}}{\log n})$, we obtain the same bound $O^*(k/m\log^2n)$ for $\abs{E_3}$, $\abs{E_5}$, $\abs{E_7}$ and $\abs{E_9}$, and conclude the proof of the claim. \qed

Combining the bounds from Claim \ref{cl_bound_E1}, \ref{cl_bound_Es} for every single term in \eqref{eq_full_decomposition_dl}, we finish the proof for Lemma \ref{lm_diagonal_entries_in_expectation}. \qedhere

\subsection{Proof of Lemma \ref{lm_reweighted_cov_matrix_in_expectation}}
\label{sec:prf_lm_reweighted_cov_matrix_in_expectation}
We prove this lemma by using the same strategy used to prove Lemma \ref{lm_diagonal_entries_in_expectation}.
\begin{align*}
  \label{eq_full_decomposition_Muv}
  M_{u,v} &\triangleq \E[\inprod{y}{u}\inprod{y}{v} y_Ry_R^T] \\
  & =
            \E[\inprod{A^*x^* + \varepsilon}{u} \inprod{A^*x^* +
            \varepsilon}{v} (\rA{R}^*x^* + \varepsilon_R)(\rA{R}^*x^*
            + \varepsilon_R)^T] \nonumber \\
  &= \E\bigl[\bigl\{\inprod{x^*}{\beta} \inprod{x^*}{\beta'} +
    x^{*T}(\beta v^T + \beta'u^T)\varepsilon +
    u^T\varepsilon\varepsilon^Tv\bigr\}\bigl\{\rA{R}^*x^*x^{*T}\rA{R}^{*T}
    + \rA{R}^*x^*\varepsilon_R^T + \varepsilon_Rx^{*T}\rA{R}^{*T} +
    \varepsilon_R\varepsilon_R^T \bigr\}\bigr] \nonumber \\
  &= M_1 + \dots + M_8,
\end{align*}
in which only nontrivial terms are kept in place, including
\begin{align}
  \begin{split}
  &M_1 = \E[\inprod{x^*}{\beta} \inprod{x^*}{\beta'}\rA{R}^*x^*x^{*T}\rA{R}^{*T}] \\
  &M_2 = \E[\inprod{x^*}{\beta}\inprod{x^*}{\beta'}\varepsilon_R\varepsilon_R^T] \\
  &M_3 = \E[x^{*T}(\beta v^T + \beta'u^T)\varepsilon\rA{R}^*x^*\varepsilon_R^T] \\
  &M_4 = \E[x^{*T}(\beta v^T + \beta'u^T)\varepsilon\varepsilon_Rx^{*T}\rA{R}^{*T}] \\
  &M_5 = \E[u^T\varepsilon\varepsilon^Tv\rA{R}^*x^*x^{*T}\rA{R}^{*T}] \\
  &M_6 = \E[u^T\varepsilon\varepsilon^Tv\rA{R}^*x^*\varepsilon_R^T] \\
  &M_7 = \E[u^T\varepsilon\varepsilon^Tv\varepsilon_R^Tx^{*T}\rA{R}^{*T}] \\
  &M_8 = \E[u^T\varepsilon\varepsilon^Tv\varepsilon_R\varepsilon_R^T]
  \end{split}
\end{align}
By swapping inner product terms and taking advantage of the independence, we can show that $M_6  =
\E[\rA{R}^*x^*u^T\varepsilon\varepsilon^Tv\varepsilon_R^T] = 0$ and
$M_7 =
\E[u^T\varepsilon\varepsilon^Tv\varepsilon_R^Tx^{*T}\rA{R}^{*T}] =
0$. The remaining are bounded in the next claims.

\begin{Claim} In the decomposition \eqref{eq_full_decomposition_Muv},
  \label{cl_bound_M1}
  \begin{align*}
  M_1 = \sum_{i \in U \cap V}q_ic_i\beta_i\beta'_i\AR{i}^*\AR{i}^{*T} + E'_1 + E'_2 + E'_3
\end{align*}
where
$E'_1 = \sum_{i \notin U \cap V}q_ic_i\beta_i\beta'_i\AR{i}^*\AR{i}^{*T}$,
$E'_2 = \sum_{i \neq j}
  q_{ij}\beta_i\beta'_i\AR{j}^*\AR{j}^{*T}$ and
$E'_3 = \sum_{i \neq j}
  q_{ij}(\beta_i\AR{i}^*\beta'_j\AR{j}^{*T} +
  \beta'_i\AR{i}^*\beta_j\AR{j}^{*T})$ have norms bounded by
  $O^*(k/m\log n)$.
\end{Claim}

\proof The expression of $M_1$ is obtained in the same way as $E_1$ is derived in the proof of Lemma \ref{lm_diagonal_entries_in_expectation}. To prove the claim, we bound all the terms with respect to the spectral norm of $\rAR^*$ and make use of Assumption \textbf{A4} to find the exact upper bound.

For the first term $E'_1$, rewrite $E'_1 = \AR{S}^*D_1\AR{S}^{*T}$ where $S = [m] \backslash (U \cap V)$ and $D_1$ is a diagonal matrix whose entries are $q_ic_i\beta_i\beta_i'$. Clearly, $\norm{D_1} \leq \max_{i \in S}\abs{q_ic_i\beta_i\beta_i'} \leq O^*(k/m\log n)$ as shown in Claim \ref{cl_bound_E1}, then
$$\norm{E'_1} \leq \max_{i \in S}\abs{q_ic_i\beta_i\beta_i'}\norm{\AR{S}^*}^2 \leq \max_{i \in S}\abs{q_ic_i\beta_i\beta_i'}\norm{\rAR^*}^2 \leq O^*(k/m\log n)$$
where $\norm{\AR{S}^*} \leq \norm{\rAR^*} \leq O(1)$. The second term $E'_2$ is a sum of positive semidefinite matrices, and $\norm{\beta} \leq O(k\log n)$, then
\begin{align*}
  E'_2 =  \sum_{i \neq j} q_{ij}\beta_i\beta'_i\AR{j}^*\AR{j}^{*T} &\preceq \max_{i\neq j}q_{ij} \Bigl( \sum_i\beta_i\beta_i' \Bigr) \Bigl( \sum_j \AR{j}^*\AR{j}^{*T} \Bigr) \preceq (\max_{i\neq j}q_{ij}) \norm{\beta} \norm{\beta'} \rAR^*\rAR^{*T}
\end{align*}
which implies that $\norm{E'_2} \leq (\max_{i\neq j}q_{ij}) \norm{\beta} \norm{\beta'} \norm{\rAR^*}^2 \leq \Otilde(k^3/m^2)$. Observe that $E_3'$ has the same form as the last term in Claim \ref{cl_bound_E1}, which is $E_3' = \rAR^{*T}Q_{\beta}\rAR^*$. Then 
$$\norm{E'_3} \leq \norm{Q_{\beta}}\norm{\rAR^*}^2 \leq (\max_{i\neq j}q_{ij}) \norm{\beta} \norm{\beta'}\norm{\rAR^*}^2 \leq \Otilde(k^3/m^2)$$

By Claim \ref{cl_bounds_of_beta}, we have $\norm{\beta}$ and
$\norm{\beta'}$ are bounded by $O(\sqrt{k}\log n)$, and note that $k
\leq O^*(\sqrt{n}/\log n)$, then we complete the proof for Lemma 6. \qed

\begin{Claim}
  \label{cl_bound_Ms}
  In the decomposition \eqref{eq_full_decomposition_Muv}, $M_2$, $M_3$, $M_4$, $M_5$ and $M_8$ have norms bounded by $O^*(k/m\log n)$.
\end{Claim}
\proof Recall the definition of $Q$ in Claim \ref{cl_bound_Es} and use the fact that $\E[x^*x^{*T}] = Q$, we can get
$M_2 = \E[\inprod{x^*}{\beta}\inprod{x^*}{\beta'}\varepsilon_R\varepsilon_R^T]
= \sum_{i}\sigmae^2q_i\beta_i\beta'_iI_r$. Then, $\norm{M_2} \leq \sigmae^2\max_{i}q_i\norm{\beta}\norm{\beta'} \leq O(\sigmae^2k^2\log^2n/m)$.

The next three terms all involve $\rA{R}^*$ whose norm is bounded according to Assumption \textbf{A4}. Specifically,
\begin{align*}
  M_3 &= \E[x^{*T}(\beta v^T + \beta'u^T)\varepsilon\rA{R}^*x^*\varepsilon_R^T] = \E[\rA{R}^*x^*x^{*T}(\beta v^T + \beta'u^T)\varepsilon\varepsilon_R^T] \\
  &= \rA{R}^*\E[x^*x^{*T}](\beta v^T + \beta'u^T)\E[\varepsilon\varepsilon_R^T] \\
  &= \rA{R}^*Q(\beta v^T + \beta'u^T)\E[\varepsilon\varepsilon_R^T],
\end{align*}
and
\begin{align*}
  M_4 &= \E[x^{*T}(\beta v^T + \beta'u^T)\varepsilon\varepsilon_Rx^{*T}\rA{R}^{*T}] = \E[\varepsilon_R\varepsilon^T(v\beta^T + u\beta'^T)x^*x^{*T}\rA{R}^{*T}] \\
  &= \E[\varepsilon_R\varepsilon^T](v\beta^T + u\beta'^T)\E[x^*x^{*T}]\rA{R}^{*T} \\
  &= \E[\varepsilon_R\varepsilon^T](v\beta^T + u\beta'^T)Q\rA{R}^{*T},
\end{align*}
and the fifth term $M_5 = \E[u^T\varepsilon\varepsilon^Tv\rA{R}^*x^*x^{*T}\rA{R}^{*T}] =
\sigmae^2u^Tv\rA{R}^*\E[x^*x^{*T}]\rA{R}^{*T} = \sigmae^2u^Tv\rA{R}^*Q\rA{R}^{*T}$. We already have $\norm{\E[\varepsilon\varepsilon_R^T]} = \sigmae^2$, $\norm{Q} \leq O(k/m)$ and $\abs{u^Tv} \leq \Otilde(k)$ (proof of Claim \ref{cl_bound_y}), then the remaining work is to bound $\norm{\beta v^T + \beta'u^T}$, then the bound of $v\beta^T + u\beta'^T$ directly follows. We have $\norm{\beta v^T}= \norm{A^*uv^T} \leq \norm{A^*}\norm{u}\norm{v} \leq \Otilde(k)$. Therefore, all three terms $M_3$, $M_4$ and $M_5$ are bounded in norm by $\Otilde(\sigmae^2k^2/m) \leq \Otilde(k^3/mn)$.

The remaining term is 
\begin{align*}
  M_8 &= \E[u^T\varepsilon\varepsilon^Tv\varepsilon_R\varepsilon_R^T] = \E[\bigl(\sum_{i, j}u_iv_j\varepsilon_i\varepsilon_j\bigr)\varepsilon_R\varepsilon_R^T] \\
      &= \E[\bigl(\sum_{i\in R}u_iv_i\varepsilon_i^2\varepsilon_R\varepsilon_R^T\bigr)] + \E[\bigl(\sum_{i \neq j}u_iv_j\varepsilon_i\varepsilon_j\bigr)\varepsilon_R\varepsilon_R^T] \\
  &= \sigmae^4u_Rv_R^T
\end{align*}
where $u_R = \rAR^*\alpha + (\varepsilon_u)_R$ and $v_R = \rAR^*\alpha' + (\varepsilon_v)_R$. We can see that $\norm{u_R} \leq \norm{\rAR^*}\norm{\alpha} + \norm{(\varepsilon_u)_R} \leq \Otilde(\sqrt{k})$. Therefore, $\norm{M8} \leq \Otilde(\sigmae^4k) = \Otilde(k^3/n^2)$. Since $m = O(n)$ and $k \leq O^*(\frac{\sqrt{n}}{\log n})$, then we can bound all the above terms by $O^*(k/m\log n)$ and finish the proof of Claim \ref{cl_bound_Ms}. \qed

Combine the results of Claim \ref{cl_bound_M1} and \ref{cl_bound_Ms}, we complete the proof of Lemma \ref{lm_reweighted_cov_matrix_in_expectation}.


%% file: appendix_algorithm.tex
\section{Analysis of Main Algorithm}
\label{appdx_main_algorithm}



\subsection{Simple Encoding}
We can see that $(A^sx - y)\sgn(x)^T$ is random over $y$ and $x$ that is obtained from the encoding step. 
We follow~\citep{arora15-neural} to derive the closed form of $g^s = \E[(A^sx - y)\sgn(x)^T]$ by proving that the encoding recovers the sign of $x^*$ with high probability as long as $A^s$ is close enough to $A^*$.

\begin{Lemma}
  \label{lm_sign_recovery_x}
  Assume that $A^s$ is $\delta$-close to $A^*$ for $\delta = O(r/n\log
  n)$ and $\mu \leq \frac{\sqrt{n}}{2k}$, and $k \geq \Omega(\log m)$ then with probability over random samples $y = A^*x^* + \varepsilon $
  \begin{equation}
    \label{eq_sign_recovery_x}
    \sgn(\thres_{C/2}{\left( (A^{s})^Ty \right)} = \sgn(x^*)
  \end{equation}
\end{Lemma}
\proof[Proof of Lemma \ref{lm_sign_recovery_x}] We follow the same proof strategy from~\citep{arora15-neural} (Lemmas 16 and 17) to prove a more general version in which the noise $\varepsilon$ is taken into account. Write $S = \supp(x^*)$ and skip the superscript $s$ on $A^s$ for the readability. What we need is to show $S = \{i \in [m] : \inprod{\cAi}{y} \geq C/2\}$ and then $\sgn(\inprod{\cAi^s}{y}) = \sgn(x_i^*)$ for each $i \in S$ with high probability.
Following the same argument of~\citep{arora15-neural},
we prove in below a stronger statement that, even conditioned on the support $S$,
$S = \{i \in [m] : \inprod{\cAi}{y} \geq C/2\}$ with high probability.

Rewrite 
$$\inprod{\cAi}{y} = \inprod{\cAi}{ A^*x^* + \varepsilon} = \inprod{\cAi}{\cAi^*}x_i^* + \sum_{j \neq i}\inprod{\cAi}{\cA{j}^*}x_j^* + \inprod{\cAi}{\varepsilon},$$
and observe that, due to the closeness of $\cAi$ and $\cAi^*$, the first term is either close to $x_i^*$ or equal to 0 depending on whether or not $i \in S$. Meanwhile, the rest are small due to the incoherence and the concentration in the weighted average of noise. We will show that both $Z_i = \sum_{S\backslash \{i\}}\inprod{\cAi}{\cA{j}^*}x_j^*$ and $\inprod{\cAi}{\varepsilon}$ are bounded by $C/8$ with high probability.

The cross-term $Z_i = \sum_{S\backslash
  \{i\}}\inprod{\cAi}{\cA{j}^*}x_j^*$ is a sum of zero-mean
independent sub-Gaussian random variables, which is another sub-Gaussian random variable with variance
$\sigma^2_{Z_i} =  \sum_{S\backslash
  \{i\}}\inprod{\cAi}{\cA{j}^*}^2$.
Note that
$$\inprod{\cAi}{\cA{j}^*}^2 \leq 2\bigl(\inprod{\cAi^*}{\cA{j}^*}^2 + \inprod{\cAi - \cAi^*}{\cA{j}^*}^2\bigr) \leq 2\mu^2/n + 2\inprod{\cAi - \cAi^*}{\cA{j}^*}^2,$$
where we use Cauchy-Schwarz inequality and the $\mu$-incoherence of $A^*$. Therefore,
$$\sigma^2_{Z_i} \leq  2\mu^2k/n + 2\norm{\cA{S}^{*T}(\cAi - \cAi^*)}_F^2 \leq 2\mu^2k/n + 2\norm{\cA{S}^*}^2\norm{\cAi - \cAi^*}^2 \leq O(1/\log n),$$
under $\mu \leq \frac{\sqrt{n}}{2k}$, to conclude $2\mu^2k/n\le O(1/\log n)$ we need $1/k=O(1/\log n)$, i.e.~$k=\Omega(\log n)$. Applying Bernstein's inequality, we get $\abs{Z_i} \leq C/8$ with high
probability. What remains is to bound the noise term
$\inprod{\cAi}{\varepsilon}$. In fact, $\inprod{\cAi}{\varepsilon}$ is
sum of $n$ Gaussian random variables, which is a sub-Gaussian with variance $\sigmae^2$.
It is easy to see that $|\inprod{\cAi}{\varepsilon}| \leq \sigmae\log n$ with high probability. 
Notice that $\sigma_\varepsilon=O(1/\sqrt{n})$.

Finally, we combine these bounds to have $\abs{Z_i +
  \inprod{\cAi}{\varepsilon}} \leq C/4$. Therefore, for $i \in S$,
then $\abs{\inprod{\cAi}{y}} \geq C/2$ and negligible otherwise. Using
union bound for every $i = 1, 2, \dots, m$, we finish the proof of the
Lemma. \qedhere

Lemma \ref{lm_sign_recovery_x} enables us to derive the expected update direction $g^s = \E[(A^sx - y)\sgn(x)^T]$ explicitly.

\subsection{Approximate Gradient in Expectation}
\proof[Proof of Lemma \ref{lm_expected_columwise_update}] Having the result from Lemma \ref{lm_sign_recovery_x}, we are now able to study the expected update direction $g^s = \E[(A^sx - y)\sgn(x)^T]$. Recall that $A^s$ is the update at the $s$-th iteration and $x \triangleq \thres_{C/2}((A^{s})^Ty)$. Based on the generative model, denote $p_i = \E[x_i^*\sgn(x_i^*) | i \in S], q_i = \Prob[i \in S]$ and $q_{ij} = \Prob[i, j \in S]$. Throughout this section, we will use
$\zeta$ to denote any vector whose norm is negligible although they can be different across their appearances. $A_{-i}$ denotes the sub-matrix of $A$ whose $i$-th column is removed. To avoid overwhelming appearance of the superscript $s$, we skip it from $A^s$ for neatness. Denote $\sgnEvent$ is the event under which the support of $x$
is the same as that of $x^*$, and $\bar{\mathcal{F}}_{x^*}$ is
its complement. In other words, $\textbf{1}_{\mathcal{F}_{x^*}} =
\1[\sgn(x)=\sgn(x^*)]$ and $\textbf{1}_{\mathcal{F}_{x^*}} + \textbf{1}_{\bar{\mathcal{F}}_{x^*}} = 1$.
\begin{align*}
  \cgi^s = \E[(Ax - y)\sgn(x_i)] &=
                                 \E[(Ax-y)\sgn(x_i)\textbf{1}_{\mathcal{F}_{x^*}}]
                                 \pm \zeta
\end{align*}
Using the fact that $y = A^*x^* + \varepsilon$ and that under
$\sgnEvent $ we have $Ax = \cAS x_S = \cAS\cAS^Ty = \cAS\cAS^TA^*x^* +
\cBS\cBS^T\varepsilon$. Using the independence of $\varepsilon$ and $x^*$ to get rid of the noise term, we get
\begin{align*}
  \cgi^s &= \E[(\cBS\cBS^T - I_n)A^*x^*\textbf{1}_{\mathcal{F}_{x^*}}] +
          \E[(\cBS\cBS^T - I_n)\varepsilon
          \sgn(x_i)\textbf{1}_{\mathcal{F}_{x^*}}]  \pm \zeta \\
        &= \E[(\cBS\cBS^T - I_n)A^*x^*
          \sgn(x_i)\textbf{1}_{\mathcal{F}_{x^*}}] \pm \zeta \quad
          \text{(Independence of $\varepsilon$ and $x$'s)} \\
        &= \E[(\cBS\cBS^T - I_n)A^*x^* \sgn(x^*_i)(1 -
          \textbf{1}_{\bar{\mathcal{F}}_{x^*}})] \pm \zeta \quad \text{(Under
          $\mathcal{F}_{x^*}$ event)} \\
        &= \E[(\cBS\cBS^T - I_n)A^*x^* \sgn(x^*_i)] \pm \zeta
\end{align*}
Recall from the generative model assumptions that $S = \supp(x^*)$ is random and the entries of $x^*$ are pairwise independent given the support, so
\begin{align*}
  \cgi^s &= \E_S\E_{x^*|S}[(\cBS\cBS^T - I_n)A^*x^* \sgn(x^*_i)] \pm \zeta \\
  &= p_i\E_{S, i \in S}[(\cBS\cBS^T - I_n)\cAi^*] \pm \zeta \\
  &= p_i\E_{S, i \in S}[(\cBi\cBi^T - I_n)\cAi^*] + p_i\E_{S, i\in S}[\sum_{l\in S, l \neq i}\cB{l}\cB{l}^T\cAi^*] \pm \zeta \\
  &= p_iq_i(\cBi\cBi^T - I_n)\cAi^* + p_i\sum_{l \in [m], l \neq i}q_{il}\cB{l}\cB{l}^T\cAi^* \pm \zeta \\
  &= p_iq_i(\lambda_i\cBi - \cAi^*) + p_i\cB{-i}\diag(q_{ij})\cB{-i}^T\cAi^* \pm \zeta
\end{align*}
where $\lambda_i^s = \langle \cAi^s, \cAi^*\rangle$.
Let $\xi_i^s =  A_{R, -i}\diag(q_{ij})\cB{-i}^T\cAi^*/q_i$ for $j = 1, \dots, m$, we now have the full expression of the expected approximate gradient at iteration $s$:
\begin{equation}
  \label{eq_expected_columwise_update}
  g_{R, i}^s = p_iq_i(\lambda_i\AR{i}^s - \AR{i}^* + \xi_i^s) \pm \zeta_{R}.
\end{equation}
What remains is to bound norms of $\xi_s$ and $\zeta$. We have $\norm{A_{R, -i}^s} \leq \norm{A_{-i}^s} \leq O(\sqrt{m/n})$ \whp\ Then, along with the fact that $\norm{A^*_i} = 1$, we can bound $\norm{\xi_i^s}$
\begin{equation}
  \norm{\xi_i^s} \leq \norm{A_{R_i, -i}^s} \max_{j \neq i} \frac{q_{ij}}{q_i} \norm{A_{-i}^s} \leq O(k/n).\label{eqn:eps_norm}
\end{equation}
Next, we show that norm of $\zeta$ is negligible. In fact, $\sgnEvent$
happens with very high probability, then it suffices to bound norm of
$(Ax - y)\sgn(x_i)$ which will be done using Lemma \ref{cl_norm_bound_ghi}
and Lemma \ref{aux_lm_pull_out_prob_bound} in Section \ref{sample_complexity}. This concludes the proof for Lemma \ref{lm_expected_columwise_update}. \qed

%% file: sample_complexity.tex
\section{Sample Complexity}
\label{sample_complexity}

In previous sections, we rigorously analyzed both initialization and
learning algorithms as if the expectations $g^s$, $e$ and $\Muv$ were
given. Here we show that corresponding estimates based on empirical
means are sufficient for the algorithms to succeed, and identify how
may samples are
required. Technically, this requires the study of their concentrations
around their expectations. Having had these concentrations, we are ready to prove Theorems \ref{main_thm_initialization} and \ref{main_thm_columnwise_descent_in_expectation}.

The entire section involves a variety of concentration bounds. Here we make heavy use of Bernstein's inequality for different types of random variables (including scalar, vector and matrix). The Bernstein's inequality is stated as follows.

\begin{Lemma} [Bernstein's Inequality]
    \label{aux_lm_bernstein}
 Suppose that $Z^{(1)}, Z^{(2)}, \dots, Z^{(p)}$ are $p$ \iid\ samples from some distribution $\mathcal{D}$. If $\E[Z] = 0$, $\norm{Z^{(j)}} \leq \Rad$ almost surely and $\norm{\E[Z^{(j)}(Z^{(j)})^T} \leq \sigma^2$ for each $j$, then 
 \begin{equation}
   \frac{1}{p}\norm[\Big]{\sum_{j=1}^p Z^{(j)}} \leq \Otilde\biggl(\frac{\Rad}{p} + \sqrt{\frac{\sigma^2}{p}}\biggr)
   \label{ineq_bernstein}
 \end{equation}
holds with probability $1-n^{-\omega(1)}$. 
\end{Lemma}

Since all random variables (or their norms) are not bounded almost surely in our model setting, we make use of a technical lemma that is used in \cite{arora15-neural} to handle the issue.

\begin{Lemma} [\cite{arora15-neural}]
  \label{aux_lm_bernstein_truncate_condition}
  Suppose a random variable $Z$ satisfies $\Prob[\norm{Z} \geq \Rad(\log(1/\rho))^C] \leq \rho$ for some constant $C > 0$, then

(a) If $p = n^{O(1)}$, it holds that $\norm{Z^{(j)}} \leq \Otilde(\Rad)$ for each $j$ with probability $1 - n^{-\omega(1)}$.

(b) $\norm{\E[Z\1_{\norm{Z} \geq \Omgtilde(\Rad)}]} = n^{-\omega(1)}$.
\end{Lemma}

This lemma suggests that if $\frac{1}{p} \sum_{i=1}^p Z^{(j)}(1 -
\1_{\norm{Z^{(j)}} \geq \Omgtilde(\Rad)})$ concentrates around its
mean with high probability, then so does $\frac{1}{p} \sum_{i=1}^p
Z^{(j)}$ because the part outside the truncation level can be
ignored. Since all random variables of our interest are sub-Gaussian
or a product of sub-Gaussian that satisfy this lemma, we can apply
Lemma \ref{aux_lm_bernstein} to the corresponding truncated random variables with carefully chosen truncation levels. Then the original random variables concentrate likewise. 

In the next proofs, we define suitable random variables and identify good bounds of $\Rad$ and $\sigma^2$ for them.  Note that in this section, the expectations are taken over $y$ by conditioning on $u$ and $v$. This aligns with the construction that the estimators of $e$ and $M_{u,v}$ are empirical averages over i.i.d. samples of $y$, while $u$ and $v$ are kept fixed.
Due to the dependency on $u$ and $v$,
these (conditional) expectations inherit randomness from $u$ and $v$, and
we will formulate probabilistic bounds for them.

The application of Bernstein's inequality requires
a bound on $\norm{\E[{Z}{Z}^T(1 - \1_{\norm{{Z}} \geq \Omgtilde(\Rad))}]}$.
We achieve that by the following technical lemma, where $\tilde{Z}$ is a
standardized version of $Z$.

\begin{Lemma}
  \label{aux_lm_pull_out_prob_bound}
  Suppose a random variable $\tilde{Z}\tilde{Z}^T = aT$ where $a \geq 0$ and $T$ is positive semi-definite.
 They are both random.
 Suppose
 $\Prob[a \geq \Aupb]=n^{-\omega(1)}$
 and $\Bupb>0$ is a constant.
 Then,
$$\norm{\E[\tilde{Z}\tilde{Z}^T(1 - \1_{\norm{\tilde{Z}} \geq \Bupb})]} \leq \Aupb\norm{\E[T]} + O( n^{-\omega(1)})$$
\end{Lemma}
\proof To show this, we make use of the decomposition $\tilde{Z}\tilde{Z}^T = aT$ and a truncation for $a$. Specifically,
\begin{align*}
 \norm{\E[\tilde{Z}\tilde{Z}^T(1 - \1_{\norm{\tilde{Z}} \geq \Bupb})]} &= \E[aT(1 - \1_{\norm{\tilde{Z}} \geq \Bupb})] \\
  &\leq \norm{\E[a(1 - \1_{a \geq \mathcal{A}})T(1 - \1_{\norm{\tilde{Z}} \geq \Bupb})]} + \norm{\E[a\1_{a \geq\mathcal{A}}T(1-\1_{\norm{\tilde{Z}} \geq \Bupb})]} \\
  &\leq \norm{\E[a(1 - \1_{a \geq \mathcal{A}})T]} + \E[a\1_{a \geq \mathcal{A}}\norm{T}(1-\1_{\norm{\tilde{Z}} \geq \Bupb})] \\
  &\leq \mathcal{A}\norm{\E[T]} + \bigl(\E[\norm{aT}^2(1-\1_{\norm{\tilde{Z}} \geq \Bupb})] \E[\1_{a \geq \mathcal{A}}]\bigr)^{1/2} \\
  &\leq \mathcal{A}\norm{\E[T]}  + \bigl(\E[\norm{\tilde{Z}}^4(1-\1_{\norm{\tilde{Z}} \geq \Bupb})]\Prob[a \geq \mathcal{A}]\bigr)^{1/2} \\
  &\leq \mathcal{A}\norm{\E[T]} + \Bupb^2\bigl(\Prob[a \geq \mathcal{A}]\bigr)^{1/2}\\ 
  &\leq \mathcal{A}\norm{\E[T]} + O( n^{-\omega(1)}),
\end{align*}
where at the third step we used $T(1 - \1_{\norm{\tilde{Z}} \geq \Bupb})] \preceq T$ because of the fact that $T$ is the positive semi-definite and $1 - \1_{\norm{\tilde{Z}} \geq \Bupb} \in \{0, 1\}$ . Then, we finish the proof of the lemma.
\qed

\subsection{Sample Complexity of Algorithm \ref{alg_neural_initialization}}
In Algorithm \ref{alg_neural_initialization}, we empirically compute
the ``scores'' $\ehat$ and the reduced weighted covariance matrix
$\Mhuv$ to produce an estimate for each column of $A^*$. Since the
construction of $\Mhuv$ depends upon the support estimate $\Rhat$
given by ranking $\ehat$, we denote it by $\Mhuv^{\Rhat}$. We will
show that we only need $p = \Otilde(m)$ samples to be able to recover
the support of one particular atom and 
up to some specified level of column-wise error with high probability.

\begin{Lemma}
  \label{lm_concentration_ehat_Mhat}
  Consider Algorithm \ref{alg_neural_initialization} in which $p$ is
  the given number of samples. For any pair $u$ and $v$, then with
  high probability a) $\norm{\ehat - e} \leq O^*(k/m\log^2n)$  when $p
  = \Omgtilde(m)$ and b) $\norm{\Mhuv^{\Rhat} - \Muv^R} \leq
  O^*(k/m\log n)$ when $p = \Omgtilde(mr)$ 
where $\Rhat$ and $R$ are respectively the estimated and correct support sets of one particular atom.
\end{Lemma}

\subsubsection{Proof of Theorem \ref{main_thm_initialization}}
\label{sec:proof-thm-initialization}

Using Lemma \ref{lm_concentration_ehat_Mhat}, we are ready to prove
the Thereom \ref{main_thm_initialization}. According to Lemma \ref{lm_diagonal_entries_in_expectation} when $U \cap V = \{i\}$, we can write $\ehat$ as
$$\ehat = q_ic_i\beta_i\beta'_i\AR{i}^* \circ \AR{i}^* + ~\text{perturbation terms} + (\ehat - e),$$
and consider $\ehat - e$ as an additional perturbation with the same magnitude $O^*(k/m\log^2n)$
in the sense of $\|\cdot\|_\infty$ \whp\ The first part of Lemma \ref{lm_support_consists_and_closeness} suggests that when $u$ and $v$ share exactly one atom $i$, then the set $\Rhat$ including $r$ largest elements of $\ehat$ is the same as $\supp(A_i^*)$ with high probability.

Once we have $\Rhat$, we again write $\Mhuv^{\Rhat}$ using Lemma \ref{lm_reweighted_cov_matrix_in_expectation} as
$$\Mhuv^{\Rhat} = q_ic_i\beta_i\beta'_i\AR{i}^*\AR{i}^{*T} + ~\text{perturbation terms} + (\Mhuv^{\Rhat} - \Muv^{R}),$$
and consider $\Mhuv^{\Rhat} - \Muv^{R}$ as an additional perturbation with the same magnitude $O^*(k/m\log n)$ in the sense of the spectral norm $\|\cdot\|$ \whp\ Using the second part of Lemma \ref{lm_support_consists_and_closeness}, we have the top singular vectors of $\Mhuv^{\Rhat}$ is $O^*(1/\log n)$ -close to $\AR{i}^*$ with high probability.

Since every vector added to the list $L$ in Algorithm~\ref{alg_neural_initialization}
is close to one of the dictionary, then $A^0$ must be $\delta$-close
to $A^*$. In addition, the nearness of$A^0$ to $A^*$ is guaranteed via
an appropriate projection onto the convex set $\mathcal{B} = \{A | A
~\text{close to } A^0 ~\text{and}~ \norm{A} \leq
2\norm{A^*} \}$. Finally, we finish the proof of Theorem \ref{main_thm_initialization}. \qedhere

\subsubsection{Proof of Lemma \ref{lm_concentration_ehat_Mhat}, Part a}
\label{sec:proof-lm_concentration_e}

For some fixed $l \in [n]$, consider $p$ \iid\ realizations $Z^{(1)}, Z^{(2)}, \dots, Z^{(p)}$ of the random variable $Z \triangleq \inprod{y}{u} \inprod{y}{v} y_l^2$, then $\ehat_l =
\frac{1}{p}\sum_{i=1}^pZ^{(i)}$ and $e_l = \E[Z]$. To show that $\norm{\ehat - e}_\infty \leq O^*(k/m\log^2n)$ holds with high probability, we first study the concentration for the $l$-th entry of $\ehat - e$ and then take the union bound over all $l = 1, 2, \dots, n$. We derive upper bounds for $\abs{Z}$ and its variance
$\E[Z^2]$ in order to apply Bernstein's inequality in \eqref{ineq_bernstein} to the truncated version of $Z$.

\begin{Claim}
  \label{cl_bound_z}
  $\abs{Z} \leq \Otilde(k)$ and $\E[Z^2] \leq \Otilde(k^2/m)$ with
  high probability.
\end{Claim}
Again, the expectation is taken over $y$ by conditioning on $u$ and $v$, and therefore is still random due to the randomness of $u$ and $v$. To show Claim \ref{cl_bound_z}, we begin with proving the following auxiliary claim.
\begin{Claim}
  \label{cl_bound_y}
  $\norm{y} \leq \Otilde(\sqrt{k})$ and $\abs{\inprod{y}{u}} \leq \Otilde(\sqrt{k})$ with high probability.
\end{Claim}
\proof From the generative model, we have 
$$\norm{y} = \norm{\cAS^*x_S^* + \varepsilon} \leq \norm{\cAS^*x_S^*} + \norm{\varepsilon} \leq \norm{\cAS^*}\norm{x_S^*} + \norm{\varepsilon}, $$
 where $S = \supp(x^*)$. From Claim \ref{cl_bound_subgaussian_rv}, $\norm{x_S^*} \leq \Otilde(\sqrt{k})$ and
$\norm{\varepsilon} \leq \Otilde(\sigmae \sqrt{n})$ \whp\ In addition, $A^*$ is overcomplete and has bounded spectral norm, then $\norm{\cAS^*}  \leq \norm{A^*} \leq O(1)$. Therefore, $\norm{y} \leq \Otilde(\sqrt{k})$ \whp,
 which is the first part of the proof. To bound the second term, we write it as
$$\abs{\inprod{y}{u}} = \abs{\inprod{\cAS^*x_S^* + \varepsilon}{u}} \leq \abs{\inprod{x_S^*}{\cAS^{*T}u}} + \abs{\inprod{\varepsilon}{u}}.$$
Similar to $y$, we have $\norm{u} \leq \Otilde(\sqrt{k})$ \whp\ and hence $\norm{\cAS^{*T}u} \leq \norm{\cAS^{*T}}\norm{u} \leq O(\sqrt{k})$ with high probability. Since $u$ and $x^*$ are independent sub-Gaussian and $\inprod{x_S^*}{\cAS^{*T}u}$ are sub-exponential with variance at most $O(\sqrt{k})$, $\abs{\inprod{x_S^*}{\cAS^{*T}u}} \leq \Otilde(k)$ \whp\ Similarly,
$\abs{\inprod{\varepsilon}{u}} \leq \Otilde(\sqrt{k})$ \whp\ Consequently,
$\abs{\inprod{y}{u}} \leq \Otilde(\sqrt{k})$ \whp,
and we conclude the proof of the claim. \qed 

\proof[Proof of Claim \ref{cl_bound_z}] We have $Z = \inprod{y}{u}\inprod{y}{v} y_l^2 = \inprod{y}{u}
\inprod{y}{v} (\inprod{\rAl^*}{x^*} + \varepsilon_l)^2$ with $\inprod{y}{u}\inprod{y}{v} \leq
\Otilde(k)$ \whp\ according to Claim \ref{cl_bound_y}. What remains is to bound $y_l^2 = (\inprod{\rAl^*}{x^*} +
\varepsilon_l)^2$. Because $\inprod{\rAl^*}{x^*}$ is sub-Gaussian with
variance $\E_S(\sum_{i \in S}A_{li}^{*2}) \leq \norm{A^{*T}}^2_{1, 2} =
O(1)$, then $\abs{\inprod{\rAl^*}{x^*}} \leq O(\log n)$ \whp\ Similarly for $\varepsilon_l$, $\abs{\varepsilon_l} \leq
O(\sigmae\log n)$ \whp\ Ultimately, $\abs{\inprod{\rAl^*}{x^*} +
\varepsilon_l} \leq O(\log n)$, and hence we obtain with high probability the bound $\abs{Z} \leq
\Otilde(k)$.

To bound the variance term, we write $Z^2 = \inprod{y}{v}^2 y_l^2\inprod{y}{u}^2 y_l^2$. Note that, from the first part, we get $\inprod{y}{v}^2 y_l^2 \leq
\Otilde(k)$ and $\abs{Z} \leq
\Otilde(k)$ \whp. We apply Lemma \ref{aux_lm_pull_out_prob_bound} with some appropriate scaling to both terms, then
$$\E[Z^2(1 - \1_{\abs{Z} \geq \Omgtilde(k)})] \leq
\Otilde(k)\E[\inprod{y}{u}^2y_l^2] + O(n^{-\omega(1)}),$$
where $\E[\inprod{y}{u}^2y_l^2]$ is equal to $e_l$ for pair $u, v$ with $v = u$.
From Lemma \ref{lm_diagonal_entries_in_expectation} and its proof in Appendix 
Section ``Analysis of Initialization Algorithm",
\begin{align*}
  \E[\inprod{y}{u}^2y_l^2] &= \sum_{i=1}^m q_ic_i\beta_i^2A^{*2}_{li} + ~\text{perturbation terms},
\end{align*}
in which the perturbation terms are bounded by $O^*(k/m\log^2n)$ \whp\ (following Claims \ref{cl_bound_E1} and
\ref{cl_bound_Es}). The dominant term $\sum_{i}q_ic_i\beta_i^2A^{*2}_{li} \leq
(\max{q_ic_i\beta_i^2})\norm{\rAl^*}^2 \leq \Otilde(k/m)$ \whp\ because $\abs{\beta_i} \leq O(\log m)$ (Claim
\ref{cl_bounds_of_beta}). Then we complete the proof of the second
part. \qed

\proof[Proof of Lemma \ref{lm_concentration_ehat_Mhat}, Part a] We are now ready to prove Part a of Lemma \ref{lm_concentration_ehat_Mhat}. We apply Bernstein's inequality in Lemma \ref{aux_lm_bernstein} for the truncated random variable $Z^{(i)}(1-\1_{\abs{Z^{(i)}} \geq \Omgtilde(\Rad)})$ with $\mathcal{R} = \Otilde(k)$ and variance $\sigma^2 = \Otilde(k^2/m)$ from Claim \ref{cl_bound_z}, then
\begin{equation}
  \norm[\bigg]{\frac{1}{p} \sum_{i=1}^p Z^{(i)}(1 - \1_{\abs{Z^{(i)}} \geq \Omgtilde(\Rad)}) - \E[Z(1 - \1_{\abs{Z} \geq \Omgtilde(\Rad)})]} \leq \frac{\Otilde(k)}{p} + \sqrt{\frac{\Otilde(k^2/m)}{p}} \leq O^*(k/m\log n),
\end{equation}
\whp\ for $p = \Omgtilde(m)$. Then $\ehat_l =
\frac{1}{p}\sum_{i=1}^pZ^{(i)}$ also concentrates with high
probability. Take the union bound over $l = 1, 2, \dots, n$, we get $
\norm{\ehat - e}_\infty \leq O^*(k/m\log n)$ with high probability and
complete the proof of \ref{lm_concentration_ehat_Mhat}, Part
a. \qedhere

\subsubsection{Proof of Lemma \ref{lm_concentration_ehat_Mhat}, Part b}
\label{sec:proof-lm_concentration_M}

Next, we will prove that $\norm{\Mhuv^{\Rhat} - \Muv^R} \leq O^*(k/m\log n)$ with
high probability. 
We only need to prove the concentration inequalities for the case when
conditioned on the event that $\Rhat$ is equivalent to $R$ \whp\ Again, what we need to derive are an upper norm bound $\mathcal{R}$ of the matrix random variable $Z \triangleq \inprod{y}{u} \inprod{y}{v} y_Ry_R^T$ and its variance.

\begin{Claim}
  \label{cl_bound_Mh_A41}
$\norm{Z} \leq \Otilde(kr)$ and $\norm{\E[ZZ^T]} \leq \Otilde(k^2r/m)$ hold with high probability.
\end{Claim}
\proof We have $\norm{Z} \leq \abs{\inprod{y}{u} \inprod{y}{v}} \norm{y_R}^2$
with $\abs{\inprod{y}{u} \inprod{y}{v}} \leq \Otilde(k)$ \whp\ (according to Claim \ref{cl_bound_y}) whereas $\norm{y_R}^2 = \sum_{i \in R}y_l^2 \leq O(r\log^2n)$ \whp\ because $y_l \leq O(\log n)$ \whp\ (proof of Claim \ref{cl_bound_z}). This implies $\norm{Z} \leq \Otilde(kr)$ \whp\ The second part is handled similarly as in the proof of Claim \ref{cl_bound_z}. We take advantage of the bounds of $\Mhuv$ in Lemma \ref{lm_reweighted_cov_matrix_in_expectation}. Specifically, using the first part $\norm{Z} \leq \Otilde(kr)$ and $\inprod{y}{v}^2\norm{y_R}^2 \leq \Otilde(kr)$, and applying Lemma \ref{aux_lm_pull_out_prob_bound}, then
$$\norm{\E[ZZ^T(1 - \1_{\norm{Z} \geq \Omgtilde(kr)})]} \leq
\Otilde(kr)\norm{\E[\inprod{y}{u}^2 y_Ry_R^T]} +
\Otilde(kr)O(n^{-\omega(1)}) \leq \Otilde(kr)\norm{M_{u, u}}, $$
where $M_{u,u}$ arises from the application of Lemma \ref{lm_reweighted_cov_matrix_in_expectation}.
Recall that
$$M_{u,u} = \sum_{i}q_ic_i\beta_i^2\AR{i}^*\AR{i}^{*T} + ~\text{perturbation terms},$$
where the perturbation terms are all bounded by $O^*(k/m\log n)$ \whp\ by Claims \ref{cl_bound_M1} and \ref{cl_bound_Ms}. In addition, $$\norm{\sum_{i}q_ic_i\beta_i^2\AR{i}^*\AR{i}^{*T}} \leq (\max_{i}q_ic_i\beta_i^2)\norm{\rAR^*}^2 \leq \Otilde(k/m)\norm{A^*}^2 \leq \Otilde(k/m)$$ \whp\ Finally, the variance bound is $\Otilde(k^2r/m)$ \whp\ \qed

Then, applying Bernstein's inequality in Lemma \ref{aux_lm_bernstein} to the truncated version of $Z$ with $\mathcal{R} = \Otilde(kr)$ and variance $\sigma^2 = \Otilde(k^2r/m)$ and obtain the concentration for the full $Z$ to get 
$$\norm{\Mhuv^R - \Muv^R} \leq \frac{\Otilde(kr)}{p} + \sqrt{\frac{\Otilde(k^2r/m)}{p}} \leq O^*(k/m\log n)$$
\whp\ when the number of samples is $p = \Omgtilde(mr)$ under \Asmp{4.1}.

We have proved that $\norm{\Mhuv^R - \Muv^R} \leq O^*(k/m\log n)$ as conditioned on
the support consistency event holds \whp\
$\norm{\Mhuv^{\Rhat} - \Muv^R} \leq O^*(k/m\log n)$ is easily followed by the law of total
probability through the tail bounds on the conditional and marginal
probabilities (i.e. $\Prob[\norm{\Mhuv^R - \Muv^R} \leq  O^*(k/m\log n)|\Rhat = R])$
and $\Prob[\Rhat \neq R]$. We finish the proof of Lemma \ref{lm_concentration_ehat_Mhat}, Part b for both cases of the spectral bounds. \qed


\subsection{Proof of Theorem \ref{main_thm_columnwise_descent_in_expectation} and Sample Complexity of Algorithm \ref{alg_neural_doubly_sdl}}

In this section, we prove Theorem \ref{main_thm_columnwise_descent_in_expectation} and identify sample complexity per iteration of Algorithm \ref{alg_neural_doubly_sdl}.
We divide the proof into two steps: 1) show that when $A^s$ is $(\delta_s, 2)$-near to $A^*$ for $\delta_s= O^*(1/\log n)$, the approximate gradient estimate $\ghat^s$ is $(\alpha, \beta, \gamma_s)$-correlated-whp with $A^*$ with $\gamma_s \leq O(k^2/mn) + \alpha o(\delta_s^2)$ 
, and 2) show that the nearness is preserved at each iteration. These correspond to showing the following lemmas:

\begin{Lemma}
  At iteration $s$ of Algorithm \ref{alg_neural_doubly_sdl}, suppose that $A^s$ has each column correctly supported and is $(\delta_s, 2)$-near to $A^*$ and that $\eta = O(m/k)$. Denote $R = \supp(\cAi^s)$, then the update $\ghat_{R, i}^s$ is $(\alpha, \beta, \gamma_s)$-correlated-whp with $\AR{i}^*$ where $\alpha = \Omega(k/m)$, $\beta = \Omega(m/k)$ and $\gamma_s \leq O(k^2/mn) + \alpha o(\delta_s^2)$ for  $\delta_s= O^*(1/\log n)$. 
  \label{lm_correlation_g_hat}
\end{Lemma}
Note that this is a finite-sample version of Lemma \ref{lm_correlation_gs}.

\begin{Lemma}
  \label{lm_nearness_finite_sample}
  If $A^s$ is $(\delta_s, 2)$-near to $A^*$ and number of samples used in step $s$ is $p=\Omgtilde(m)$, then with high probability $\norm{A^{s+1} - A^*} \leq 2\norm{A^*}$.
\end{Lemma}


\proof[Proof of Theorem \ref{main_thm_columnwise_descent_in_expectation}] The correlation of $\ghat_i$ with $A_i^*$, described in Lemma \ref{lm_correlation_g_hat}, implies the descent of column-wise error according to Theorem \ref{thm_descent_from_correlation_z}. Along with Lemma \ref{lm_nearness_finite_sample}, the theorem follows directly.

\subsubsection{Proof of Lemma \ref{lm_correlation_g_hat}}

We prove Lemma \ref{lm_correlation_g_hat} by obtaining a tail bound on the difference between $\ghat^s_{R, i}$ and $\gR{i}^s$ using the Bernstein's inequality in Lemma \ref{aux_lm_bernstein}.

\begin{Lemma} 
At iteration $s$ of Algorithm \ref{alg_neural_doubly_sdl}, suppose
that $A^s$ has each column correctly supported and is $(\delta_s,
2)$-near to $A^*$. For $R = \supp(A_i^s) = \supp(A_i^*)$, then
$\norm{\ghat^s_{R, i} - \gR{i}^s} \leq O(k/m)\cdot( o(\delta_s) +
O(\epsilon_s))$ with high probability for $\delta_s = O^*(1/\log n)$ and $\epsilon_s = O(\sqrt{k/n})$ when $p = \Omgtilde(m + \sigmae^2\frac{mnr}{k})$.
\label{lm_concentration_of_gradient_sparse_case}
\end{Lemma}

To prove this lemma, we study the concentration of $\ghat^s_{R, i}$,
which is a sum of random vector of the form $(y - Ax)_R\sgn(x_i)$. We
consider random variable $Z \triangleq (y - Ax)_R\sgn(x_i) | i \in S$,
with $S = \supp(x^*)$ and $x = \thres_{C/2}(A^Ty)$. Then, using the following technical lemma to bridge the gap in concentration of the two variables. We adopt this strategy from \cite{arora15-neural} for our purpose.

\begin{Claim}
  \label{cl_concentration_sparse_Zr}
  Suppose that $Z^{(1)}, Z^{(2)}, \dots, Z^{(N)}$ are \iid\ samples of the random variable $Z = (y - Ax)_R\sgn(x_i) | i \in S$. Then,
 \begin{equation}
   \norm[\Big]{\frac{1}{N}\sum_{j=1}^N Z^{(j)} - \E[Z]} \leq o(\delta_s) + O(\epsilon_s)
   \label{ineq_bernstein}
 \end{equation}
holds with probability when $N = \Omgtilde(k + \sigmae^2nr)$,
$\delta_s = O^*(1/\log n)$ and $\epsilon_s = O(\sqrt{k/n})$. 
 \end{Claim}

\proof[Proof of Lemma \ref{lm_concentration_of_gradient_sparse_case}] Once we have done the proof of Claim \ref{cl_concentration_sparse_Zr}, we can easily prove Lemma \ref{lm_concentration_of_gradient_sparse_case}. We recycle the proof of Lemma 43 in \cite{arora15-neural}.

Write $W = \{j: i \in \supp(x^{*(j)})\}$ and $N = |W|$, then express $\ghat_{R,i}$ as 
$$\ghat_{R,i} = \frac{N}{p}\frac{1}{N} \sum_{j}(y^{(j)} - Ax^{(j)})_R\sgn(x_i^{(j)}),$$
where $\frac{1}{|W|} \sum_{j}(y^{(j)} - Ax^{(j)})_R\sgn(x_i^{(j)})$ is distributed as $\frac{1}{N}\sum_{j=1}^N Z^{(j)}$ with $N = |W|$. Note that $\E[(y - Ax)_R\sgn(x_i)] = \E[(y - Ax)_R\sgn(x_i)\1_{i\in S}] = \E[Z]\Prob[i \in S] = q_i\E[Z]$ with $q_i = \Theta(k/m)$. Following Claim \ref{cl_concentration_sparse_Zr}, we have 

$$\norm{\ghat^s_{R, i} - \gR{i}^s} \leq O(k/m)\norm[\Big]{\frac{1}{N}\sum_{j=1}^N Z^{(j)} - \E[Z]} \leq  O(k/m)\cdot( o(\delta_s) + O(\epsilon_s)),$$
holds with high probability as $p = \Omega(mN/k)$. Substituting $N$ in Claim \ref{cl_concentration_sparse_Zr}, we obtain the results in Lemma \ref{lm_concentration_of_gradient_sparse_case}. \qedhere

\proof[Proof of Claim \ref{cl_concentration_sparse_Zr}] We are now ready to prove the claim. What we need are good bounds for $\norm{Z}$ and its variance, then we can apply Bernstein's inequality in Lemma \ref{aux_lm_bernstein} for the truncated version of $Z$, then $Z$ is also concentrates likewise.

\begin{Claim}
  \label{cl_norm_bound_ghi}
  $\norm{Z} \leq \Rad$ holds with high probability for $\Rad = \Otilde(\delta_s\sqrt{k} + \mu k/\sqrt{n} + \sigmae \sqrt{r})$ with $\delta_s = O^*(1/\log n)$.
\end{Claim}
\proof From the generative model and the support consistency of the encoding step, we have $y = A^*x^* + \varepsilon = \cAS^*x^*_S + \varepsilon$ and $x_S = \cAS^Ty = \cAS^T\cAS^*x^*_S + \cAS^T\varepsilon$. Then,
\begin{align*}
   (y - Ax)_R &= (\AR{S}^*x^*_S + \varepsilon_R) - \AR{S}\cAS^T\cAS^*x^*_S - \AR{S}\cAS^T\varepsilon \\
   &= (\AR{S}^* - \AR{S})x^*_S + \AR{S}(I_k - \cAS^T\cAS^*)x^*_S + (I_n - \cAS\cAS^T)_{R\bigcdot}\varepsilon.
\end{align*}
Using the fact that $x^*_S$ and $\varepsilon$ are sub-Gaussian and that $\norm{Mw} \leq \Otilde(\sigma_w\norm{M}_F)$ holds with high probability for a fixed $M$ and a sub-Gaussian $w$ of variance $\sigma_w^2$, we have
$$\norm{(y - Ax)_R\sgn(x_i)} \leq \Otilde(\norm{\AR{S}^* - \AR{S}}_F + \norm{\AR{S}(I_k - \cAS^T\cAS^*)}_F + \sigmae\norm{(I_n - \cAS\cAS^T)_{R\bigcdot}}_F).$$
Now, we need to bound those Frobenius norms. The first quantity is easily bounded as 
\begin{equation}
  \label{ineq_C2.1}
  \norm{\AR{S}^* - \AR{S}}_F \leq \norm{\cAS^* - \cAS}_F \leq \delta_s\sqrt{k},
\end{equation}
 since $A$ is $\delta_s$-close to $A^*$. To handle the other two, we use the fact that $\norm{UV}_F \leq \norm{U} \norm{V}_F$. Using this fact for the second term, we have
$$\norm{\AR{S}(I_k - \cAS^T\cAS^*)}_F \leq \norm{\AR{S}}\norm{(I_k - \cAS^T\cAS^*)}_F, $$
where $\norm{\AR{S}} \leq \norm{\rAR} \leq O(1)$ due to the nearness.
The second part is rearranged to take advantage of the closeness and incoherence properties:
\begin{align*}
  \label{ineq_C2.2}
   \norm{I_k - \cAS^T\cAS^*}_F  &\leq \norm{I_k - \cAS^{*T}\cAS^* - (\cAS - \cAS^*)^T\cAS^*}_F \\
        &\leq \norm{I_k - \cAS^{*T}\cAS^*}_F + \norm{(\cAS - \cAS^*)^T\cAS^*}_F \\
  &\leq \norm{I_k - \cAS^{*T}\cAS^*}_F + \norm{\cAS^*}\norm{\cAS - \cAS^*}_F \\
        &\leq \mu k/\sqrt{n} + O(\delta_s\sqrt{k}),
\end{align*}
where we have used $\norm{I_k - \cAS^{*T}\cAS^*}_F \leq \mu k/\sqrt{n}$ because of the $\mu$-incoherence of $A^*$,  $\norm{\cAS - \cAS^*}_F \leq \delta_s\sqrt{k}$ in \eqref{ineq_C2.1} and $\norm{\cAS^*} \leq \norm{A^*} \leq O(1)$. Accordingly, the second Frobenius norm is bounded by 
\begin{equation}
  \label{ineq_C2.31}
  \norm{\AR{S}(I_k - \cAS^T\cAS^*)}_F \leq  O\bigl(\mu k/\sqrt{n} + \delta_s\sqrt{k}\bigr).
\end{equation}
%
The noise term is handled using the eigen-decomposition $U\Lambda U^T$ of $\cAS\cAS^T$, then with high probability
\begin{equation}
  \label{ineq_C2.4}
  \norm{(I_n - \cAS\cAS^T)_{R\bigcdot}}_F  = \norm{(UU^T - U\Lambda U^T)_{R\bigcdot}}_F = \norm{U_{R\bigcdot}(I_n - \Lambda)}_F \leq \norm{I_n - \Lambda}\norm{U_{R\bigcdot}}_F \leq O(\sqrt{r}),
\end{equation}
where the last inequality $\norm{I_n - \Lambda} \leq O(1)$ follows by
$\norm{\cAS} \leq \norm{A} \leq \norm{A - A^*} + \norm{A^*} \leq
3\norm{A^*} \leq O(1)$ due to the nearness. Putting
\eqref{ineq_C2.1}, \eqref{ineq_C2.31}
and
\eqref{ineq_C2.4} together, we obtain the bounds in Claim \ref{cl_norm_bound_ghi}. \qed
 
Next, we determine a bound for the variance of $Z$.

\begin{Claim} 
  \label{cl_bound_variance_ghi}
$\E[\norm{Z}^2] = \E[\norm{(y-Ax)_R\sgn(x_i)}^2|i\in S] \leq \sigma^2$  holds with high probability for $\sigma^2 = O(\delta_s^2k + k^2/n + \sigmae^2r)$ with $\delta_s = O^*(1/\log n)$.
\end{Claim}
\proof We explicitly calculate the variance using the fact that $x_S^*$ is conditionally independent given $S$, and so is $\varepsilon$. $x_S^*$ and $\varepsilon$ are also independent and have zero mean. Then we can decompose the norm into three terms in which the dot product is zero in expectation and the others can be shortened using the fact that $E[x^*_Sx^{*T}_S] = I_k$, $E[\varepsilon\varepsilon^T] = \sigmae I_n$.
\begin{align*}
\E[\norm{(y-Ax)_R\sgn(x_i)}^2|i\in S] &= \E[\norm{(\AR{S}^* - \AR{S}\cAS^T\cAS^*)x^*_S + (I_n - \cAS\cAS^T)_{R\cdot}\varepsilon}^2|i\in S]] \\
&= \E[\norm{\AR{S}^* - \AR{S}\cAS^T\cAS^*}_F^2| i\in S] + \sigmae^2\E[\norm{I_n - \cAS\cAS^T)_{R\bigcdot}}_F^2|i \in S].
\end{align*}
Then, by  re-writing $\AR{S}^* - \AR{S}\cAS^T\cAS^*$ as before, we get the form $(\AR{S}^* - \AR{S}) + \AR{S}(I_k - \cAS^T\cAS^*)$ in which the first term has norm bounded by $\delta_s\sqrt{k}$. The second is further decomposed as 
\begin{align}
  \label{ineq_C2.5}
  \E[\norm{\AR{S}(I_k - \cAS^T\cAS^*)}_F^2|i \in S] &\leq \sup_S\norm{\AR{S}}^2\E[\norm{I_k - \cAS^T\cAS^*}_F^2|i \in S],
\end{align}
where $\sup_S\norm{\AR{S}} \leq \norm{\rAR} \leq O(1)$. 
We will bound $\E[\norm{I_k - \cAS^T\cAS^*}_F^2|i \in S] \leq O(k\delta_s^2) + O(k^2/n)$ using the proof from \cite{arora15-neural}:
\begin{align*}
  &\E[\norm{I_k - \cAS^T\cAS^*}_F^2|i \in S] = \E[\sum_{j \in S}(1 - \cAj^T\cAj^*)^2 + \sum_{j \in S}\norm{\cAj^TA_{\bigcdot, -j}^*}^2|i \in S] \\
  &= \E[\sum_{j \in S}\frac{1}{4}\norm{\cAj - \cAj^*}^2] +  q_{ij}\sum_{j \neq i}\norm{\cAj^TA_{\bigcdot, -j}^*}^2 + q_i \norm{\cAi^TA_{\bigcdot, -i}^*}^2 + q_i \norm{A_{\bigcdot, -i}^T\cAi^*}^2,
\end{align*}
where $A_{\bigcdot, -i}$ is the matrix $A$ with the $i$-th column removed, $q_{ij} \leq O(k^2/m^2)$ and $q_i \leq O(k/m)$. For any $j = 1, 2, \dots, m$,
\begin{align*}
  \norm{\cAj^TA_{\bigcdot, -j}^*}^2 &= \norm{\cAj^{*^T}A_{\bigcdot, -j}^* + (\cAj - \cAj^*)^TA_{\bigcdot, -j}^*}^2 \\
&\leq \sum_{l \neq j}\inprod{\cAj^*}{\cAl^*}^2 + \norm{(\cAj - \cAj^*)^TA_{\bigcdot, -j}^*}^2 \\
  &\leq \sum_{l \neq j}\inprod{\cAj^*}{\cAl^*}^2 + \norm{\cAj - \cAj^*}^2\norm{A_{\bigcdot, -j}^*}^2 \leq \mu^2 + \delta_s^2.
\end{align*}
The last inequality invokes the $\mu$-incoherence, $\delta$-closeness and the spectral norm of $A^*$. Similarly, 
we come up with the same bound for $\norm{\cAi^TA_{\bigcdot, -i}^*}^2$ and $\norm{A_{\bigcdot, -i}^T\cAi^*}^2$. Consequently,
\begin{align}
  \label{ineq_C2.6}
  \E[\norm{I_k - \cAS^T\cAS^*}_F^2|i \in S] \leq O(k\delta_s^2) + O(k^2/n).
\end{align}

For the last term, we invoke the inequality \eqref{ineq_C2.4} (Claim \ref{cl_norm_bound_ghi}) to get
\begin{equation}
  \label{ineq_C2.7}
  \E[\norm{(I_n - \cAS\cAS^T)_{R\bigcdot}}_F^2|i \in S] \leq r
\end{equation}
Putting \eqref{ineq_C2.5}, \eqref{ineq_C2.6} and \eqref{ineq_C2.7}
together and using $\norm{\rAR} \leq 1$, we obtain the variance bound of $Z$: $\sigma^2 = O(\delta_s^2k + k^2/n + \sigmae^2r)$ with $\delta_s = O(1/\log^2n)$ 
. Finally, we complete the proof. \qedhere

We now apply truncated Bernstein's inequality to the random variable $Z^{(j)}(1-1_{\norm{Z^{(j)}} \geq \Omega(\Rad)})$ with $\Rad$ and $\sigma^2$ in Claims \ref{cl_norm_bound_ghi} and \ref{cl_bound_variance_ghi}, which are $\Rad = \Otilde(\delta_s\sqrt{k} + \mu k/\sqrt{n} + \sigmae \sqrt{r})$ and  $\sigma^2 = O(\delta_s^2k + k^2/n + \sigmae^2r)$. Then, $(1/N)\sum_{j=}^N Z^{(j)}$ also concentrates:
\begin{equation*}
  \norm[\Big]{\frac{1}{N} \sum_{i=1}^N Z^{(j)} - E[Z]} \leq \Otilde\Bigl(\frac{\Rad}{N}\Bigr) + \Otilde\biggl(\sqrt{\frac{\sigma^2}{N}}\biggr) = o(\delta_s) + O(\sqrt{k/n})
\end{equation*}
holds with high probability when $N = \Omgtilde(k + \sigmae^2nr)$.
%
Then, we finally finish the proof of Claim \ref{cl_concentration_sparse_Zr}. \qed

\proof[Proof of Lemma \ref{lm_correlation_g_hat}] With Claim \ref{cl_concentration_sparse_Zr}, we study the concentration of $\ghat^s_{R, i}$ around its mean $g^s_{R, i}$. Now, we consider this difference as an error term of the expectation $g^s_{R, i}$ and using Lemma \ref{lm_correlation_gs} to show the correlation of $\ghat^s_{R, i}$. Using the expression in Lemma \ref{lm_expected_columwise_update} with high probability, we can write

$$\ghat^s_{R, i} = g^s_{R, i} + (g^s_{R, i} - \ghat^s_{R, i}) = 2\alpha(\AR{i} - \AR{i}^*) + v,$$
where $\norm{v} \leq \alpha \norm{\AR{i} - \AR{i}^*} + O(k/m)\cdot( o(\delta_s) + O(\epsilon_s))$. By Lemma \ref{lm_correlation_gs}, we have $\ghat^s_{R, i}$ is $(\alpha, \beta, \gamma_s)$-correlated-whp with $\AR{i}^*$ where $\alpha = \Omega(k/m)$, $\beta = \Omega(m/k)$ and $\gamma_s \leq O(k/m)\cdot( o(\delta_s) + O(\sqrt{k/n}))$ 
, then we have done the proof Lemma \ref{lm_correlation_g_hat}. \qedhere

\subsubsection{Proof of Lemma \ref{lm_nearness_finite_sample}}
\label{sec:proof_lm_15}

We have shown the correlation of $\ghat^s$ with $A^*$ \whp\ and established the descent property of Algorithm \ref{alg_neural_doubly_sdl}. The next step is to show that the nearness is preserved at each iteration. To prove $\norm{A^{s+1} - A^*} \leq 2\norm{A^*}$ holds with high probability, we recall the update rule 
$$A^{s+1} = A^s - \eta \mathcal{P}_H(\ghat^s),$$
 where $\mathcal{P}_H(\ghat^s) = H \circ \ghat^s$. Here $H = (h_{ij})$
 where $h_{ij} = 1$ if $i \in \supp(\cAj)$ and $h_{ij} = 0$
 otherwise. Also, note that $A^s$ is $(\delta_s, 2)$-near to $A^*$ for
 $\delta_s = O^*(1/\log n)$. We already proved that this holds for the
 exact expectation $g^s$ in Lemma
 \ref{lm_nearness_infinite_sample}. To prove for $\ghat^s$, we again
 apply matrix Bernstein's inequality to bound
 $\norm{\mathcal{P}_H(g^s) - \mathcal{P}_H(\ghat^s)}$ by $O(k/m)$
 because $\eta = \Theta(m/k)$ and $\norm{A^*} = O(1)$. 

Consider a matrix random variable $Z \triangleq \mathcal{P}_H((y - Ax)\sgn(x)^T)$. Our goal is to bound the spectral norm $\norm{Z}$ and, both $\norm{\E[ZZ^T]}$ and $\norm{\E[Z^TZ]}$ since $Z$ is asymmetric. To simplify our notations, we denote by $x_R$ the vector $x$ by zeroing out the elements not in $R$. Also, denote $R_i = \supp(h_i)$ and $S = \supp(x)$. Then $Z$ can be written explicitly as
$$Z = [(y - Ax)_{R_1}\sgn(x_1), \dots, (y - Ax)_{R_m}\sgn(x_m)],$$
where many columns are zero since $x$ is $k$-sparse. The following claims follow from the proof of Claim 42 in \cite{arora15-neural}. Here we state and detail some important steps.

\begin{Claim}
  \label{cl_norm_bound_gh}
  $\norm{Z} \leq \Otilde(k)$ holds with high probability.
\end{Claim}
\proof With high probability 
$$\norm{Z} \leq  \sqrt{\sum_{i \in S}\norm{(y - Ax)_{R_i}\sgn(x_i)}^2} \leq \sqrt{k}\norm{(y - Ax)_{R_i}}$$
where we use Claim \ref{cl_norm_bound_ghi} with $\norm{(y - Ax)_R}
\leq \Otilde(\delta_s\sqrt{k})$ \whp, then $\norm{Z} \leq \Otilde(k)$
holds \whp \qed

\begin{Claim}
  \label{cl_variance_bound_gh}
  $\norm{\E[ZZ^T]} \leq O(k^2/n)$ and $\norm{\E[Z^TZ]} \leq \Otilde(k^2/n)$ with high probability.
\end{Claim}
\proof The first term is easily handled. Specifically, with high probability
$$\norm{\E[ZZ^T]} \leq \norm{\E[\sum_{i \in S} (y - Ax)_{R_i}\sgn(x_i)^2(y - Ax)^T_{R_i}]} = \norm{\E[\sum_{i \in S}(y - Ax)_{R_i}(y - Ax)_{R_i}^T]} \leq O(k^2/n),$$
where the last inequality follows from the proof of Claim 42 in
\cite{arora15-neural}, which is tedious to be repeated.



To bound $\norm{\E[Z^TZ]}$, we use bound of the full matrix
$(y-Ax)\sgn(x)^T$. Note that $\norm{y - Ax} \leq \Otilde(\sqrt{k})$
\whp\ is similar to what derived in Claim \ref{cl_norm_bound_ghi}. Then
with high probability,
$$\norm{\E[Z^TZ]} \leq \norm{\E[\sgn(x)(y - Ax)^T(y - Ax)\sgn(x)^T]} \leq  \Otilde(k)\norm{\E[\sgn(x)\sgn(x)^T]} \leq \Otilde(k^2/m).$$
where $\E[\sgn(x)\sgn(x)^T] = \diag(q_1, q_2, \dots, q_m)$ has norm bounded by $O(k/m)$. We now can apply Bernstein's inequality for the truncated version of $Z$ with $\Rad= \Otilde(k)$ and $\sigma^2 = \Otilde(k^2/m)$, then with $p = \Otilde(m)$,
$$\norm{\mathcal{P}_H(g^s) - \mathcal{P}_H(\ghat^s)} \leq \frac{\Otilde(k)}{p} + \sqrt{\frac{\Otilde(k^2/m)}{p}} \leq O^*(k/m)$$
holds with high probability. Finally, we invoke the bound $\eta = O(m/k)$ and complete the proof. \qed


%% file: appendix_ortho_dictionary.tex
\section{A Special Case: Orthonormal $A^*$}
\label{orthonormal_case}

We extend our results for the special case where the dictionary is orthonormal. As such, the dictionary is perfectly incoherent and bounded (i.e., $\mu = 0$ and $\norm{A^*} = 1$).

\begin{Theorem}
\label{thm_initialization_orthonormal}
  Suppose that $A^*$ is orthonormal. When $p_1 = \Omgtilde(n)$ and $p_2 =
  \Omgtilde(nr)$, then with high probability Algorithm \ref{alg_neural_initialization} returns an initial estimate $A^0$ whose columns share the same support as $A^*$ and with $(\delta, 2)$-nearness to $A^*$ with $\delta = O^*(1/\log n)$. The sparsity of $A^*$ can be achieved up to $r = O^*\bigl( \min(\frac{\sqrt{n}}{\log^2 n}, \frac{n}{k^2\log^2n}) \bigr)$.
\end{Theorem}

We use the same initialization procedure for this special case and achieve a better order of $r$.
The proof of Theorem~\ref{thm_initialization_orthonormal} follows the analysis for the general case with following two results:

\begin{Claim}[Special case of Claim \ref{cl_bounds_of_beta}]
   \label{cl_bounds_of_beta_spec}
  Suppose that $u = A^*\alpha + \varepsilon_u$ is a random sample and $U = \supp(\alpha)$. Let $\beta = A^{*T}u$, then \whp,
   we have (a) $\abs{\beta_i - \alpha_i} \leq \sigmae\log n$ for each $i$  and
  (b) $\norm{\beta} \leq O(\sqrt{k}\log n + \sigmae\sqrt{n}\log n)$.
\end{Claim}
\proof We have $\beta = A^{*T}u = \alpha + A^{*T}\epsilon_u$, then $\beta_i - \alpha_i = \inprod{\cAi^*}{\epsilon_u}$ and $\norm{\beta - \alpha} = \norm{\epsilon_u}$. Using probability bounds of $\inprod{\cAi^*}{\epsilon_u}$, $\norm{\epsilon_u}$ and $\norm{\alpha}$ in Claim~\ref{cl_bound_subgaussian_rv}, we have the claim proved.~\qedhere

We draw from the claim that for any $i \notin U \cap V$, $\abs{\beta_i\beta'_i} \leq O(\sigmae\log^2n)$ and have the following result: 

\begin{Lemma} 
Fix samples $u$ and $v$ and suppose that $y = A^*x^* + \varepsilon$ is a random sample independent of $u,v$. 
The expected value of the score for the $l^{\textrm{th}}$ component of $y$ is given by:
  \begin{align*}
    e_l \triangleq  \E[\inprod{y}{u}\inprod{y}{v} y_l^2] 
    = \sum_{i \in U \cap V}q_ic_i\beta_i\beta'_iA^{*2}_{li} + ~\text{perturbation terms}
  \end{align*}
where $q_i = \Prob[i \in S]$, $q_{ij} = \Prob[i, j \in S]$ and $c_i =
\E[x_i^4|i \in S]$. Moreover, the perturbation terms have absolute value at most $O^*\bigl(k/n\log^2n \max (1/\sqrt{n}, k^2/n) \bigr) $.
\label{lm_diagonal_entries_in_expectation_spec}
\end{Lemma}

\proof Lemma follows Lemma \ref{lm_diagonal_entries_in_expectation} via Claim \ref{cl_bounds_of_beta} except that the second term of $E_1$ is bounded by $O(k\log^2n/n^{3/2})$.


%% file: appendix_more_arora.tex
\section{Extensions of \citet{arora15-neural}}
\label{appdx_improve_arora}

\subsection{Sample complexity in noisy case}
In this section, we study the sample complexity of the algorithms in~\citet{arora15-neural} in the presence of noise. While noise with order $\sigmae = O(1/\sqrt{n})$ does not change the sample complexity of the initialization algorithm, it affects that of the descent stage. The analysis involves producing a sharp bound for $\norm{\ghat^s_{\bigcdot, i} - \cgi^s}$.

\begin{Lemma} 
For a regular dictionary $A^*$, suppose $A^s$ is $(\delta_s, 2)$-near to $A^*$ with $\delta_s = O^*(1/\log n)$, then with high probability $\norm{\ghat^s_{\bigcdot, i} - \cgi^s} \leq O(k/m)\cdot( o(\delta) + O(\sqrt{k/n}))$ when $p = \Omgtilde(m + \sigmae^2\frac{mn^2}{k})$.
\label{lm_concentration_of_gradient_regular_case}
\end{Lemma}
\proof This follows directly from Lemma~\ref{lm_concentration_of_gradient_sparse_case} where $r = n$.~\qedhere

We tighten the original analysis to obtain the complexity $\Omgtilde(m)$ instead of $\Omgtilde(mk)$ for the noiseless case. Putting together with $p = \Omgtilde(mk)$ required by the initialization, we then have the overall sample complexity $\Otilde(mk + \sigmae^2\frac{mn^2}{k})$ for the algorithms in~\citet{arora15-neural} in the noise regime.

\subsection{Extension of~\citet{arora15-neural}'s initialization algorithm for sparse case}
We study a simple and straightforward extension of the initialization algorithm of~\citet{arora15-neural} for the sparse case.
This extension is produced by adding an extra projection, and is described in Figure~\ref{alg_arora_modified}. The recovery of the support of $A^*$ is guaranteed by the following Lemma:

\begin{Lemma} Suppose that $z^* \in \R^n$ is $r$-sparse whose nonzero entries are at least $\tau$ in magnitude. Provided $z$ is $\delta$-close to $z^*$ and $z^0 = \mathcal{H}_r(z)$ with $\delta = O^*(1/\log n)$ and $r = O^*(\log^2 n)$, then $z^0$ and $z^*$ has the same support.
\end{Lemma}
\proof Since $z^0$ is $\delta$-close to $z^*$, then $\Vert z^0 - z^*\Vert \leq \delta$ and $|z_i - z_i^*| \leq \delta$ for every $i$.
For $i \in \supp(z^*)$, $$|z_i| \geq |z_i^*| - |z_i - z_i^*| \geq \tau - \delta$$
 and for $i \notin \supp(z^*)$, $|z_i| \leq \delta$. Since $\tau > O(1/\sqrt{r}) \gg \delta $, then the $r$-largest entries of $z$ are in the support $z^*$, and hence $z^0$ and $z^*$ has the same support. \qedhere

\begin{algorithm}[!h] 
  \begin{algorithmic} 
    \State \textbf{Initialize} $L = \emptyset$
    \State Randomly divide $p$ samples into two disjoint sets $\mathcal{P}_1$ and $\mathcal{P}_2$ of sizes $p_1$ and $p_2$ respectively
    \State \textbf{While} $|L| < m$. Pick $u$ and $v$ from $\mathcal{P}_1$ at random
    \State \indent Reconstruct the re-weighted covariance matrix $\Mhuv$:
    $$\Mhuv = \frac{1}{p_2}\sum_{i=1}^{p_2}\langle y^{(i)}, u\rangle \langle y^{(i)}, v\rangle y^{(i)}(y^{(i)})^T$$
    \State \indent Compute the top singular values $\delta_1, \delta_2$ and top singular vector $z$ of $\Mhuv$
    \State \indent \textbf{If} $\delta_1 \geq \Omega(k/m)$ and $\delta_2 < O^*(k/m\log n)$
    \State \indent \indent $z = \mathcal{H}_r(z)$, where $\mathcal{H}_r$ keeps $r$ largest entries of $z$
    \State \indent \indent \textbf{If} $z$ is not within distance $1/\log n$ of any vector in $L$ even with sign flip
    \State \indent \indent \indent $L = L \cup \{z\}$
    \State \textbf{Return} $A^0 = (L_1, \dots, L_m)$
  \end{algorithmic}
\caption{Pairwise Reweighting with Hard-Thresholding}
\label{alg_arora_modified}
\end{algorithm}

\begin{Theorem}
\label{main_initialization_ext}
  Suppose that Assumptions \textnormal{\textbf{B1-B4}} hold and
  Assumptions \textnormal{\textbf{A1-A3}} satify with $\mu =
  O^*\bigl(\frac{\sqrt{n}}{k\log^3n}\bigr)$
  and $r = O^*(\log^2 n)$. When $p_1 = \Omgtilde(m)$ and $p_2 =
  \Omgtilde(mk)$, then with high probability Algorithm \ref{alg_arora_modified} returns an initial estimate $A^0$ whose columns share the same support as $A^*$ and with $(\delta, 2)$-nearness to $A^*$ with $\delta = O^*(1/\log n)$.
\end{Theorem}
This algorithm requires $r = O^*(\log^2n)$, which is somewhat better
than ours. However, the sample complexity and running time is inferior
as compared with our novel algorithm.


%% file: appendix_neural_impl.tex
\section{Neural Implementation of Our Approach}
\label{neural_implementation}

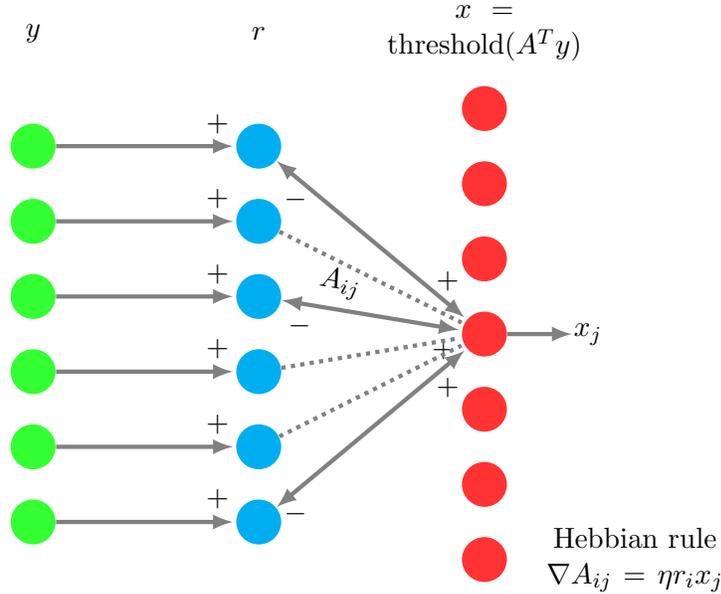
\begin{figure}[h]
  \centering
  \input{neural_diag}
  \caption{Neural network implementation of Algorithm~\ref{alg_neural_doubly_sdl}. The network takes the image $y$ as input and produces the sparse representation $x$ as output. The hidden layer represents the residual between the image and its reconstruction $Ax$. The weights $A_{ij}$'s are stored on synapses, but most of them are zero and shown by the dotted lines.} 
\label{neural_diag}
\end{figure}

We now briefly describe why our algorithm is ``neurally plausible''. Basically, similar to the argument  in~\cite{arora15-neural}, we describe at a very high level how our algorithm can be implemented via a neural network architecture. 
One should note that although both our initialization and descent stages are non-trivial modifications of those in~\cite{arora15-neural}, both still inherit the nice neural plausiblity property. 

\subsection{Neural implementation of Stage 1: Initialization}

Recall that the initialization stage includes two main steps: (i) estimate the support of each column of the synthesis matrix, and (ii) compute the top principal component(s) of a certain truncated weighted covariance matrix. Both steps involve simple vector and matrix-vector manipulations that can be implemented plausibly using basic neuronal manipulations. 

For the support estimation step, we compute the product
$\inprod{y}{u}\inprod{y}{u} y \circ y$, followed by a thresholding.  The inner products, $\inprod{y}{u}$ and $\inprod{y}{v}$ can be computed using neurons via an online manner where the samples arrive in sequence; the thresholding can be implemented via a ReLU-type non-linearity. 

For the second step, it is well known that the top principal components of a matrix can be computed in a neural (Hebbian) fashion using Oja's Rule~\cite{oja}. 

\subsection{Neural implementation of Stage 2: Descent} 

Our neural implementation of the descent stage
(Algorithm~\ref{alg_neural_doubly_sdl}), shown in Figure~\ref{neural_diag}, mimics the architecture of
\cite{arora15-neural}, which describes a simple two-layer network
architecture for computing a single gradient update of $A$. The only
difference in our case is that most of the value in $A$ are set to
zero, or in other words, our network is sparse. The network takes
values $y$ from the input layer and produce $x$ as the output; there
is an intermediate layer in between connecting the middle layer with
the output via synapses. The synaptic weights are stored on $A$. The weights are updated by Hebbian learning. In our case, since $A$ is sparse (with support given by $R$, as estimated in the first stage), we enforce the condition the corresponding synapses are inactive. In the output layer, as in the initialization stage, the neurons can use a ReLU-type non-linear activation function to enforce the sparsity of $x$.


%% file: neural_diag.tex
\def\layersep{3cm}
\def\insize{6}
\def\outsize{7}

\begin{tikzpicture}[shorten >=1pt,->,draw=black!50, node distance=\layersep]
    \tikzstyle{every pin edge}=[<-,shorten <=1pt]
    \tikzstyle{neuron}=[circle,fill=black!25,minimum size=17pt,inner sep=0pt]
    \tikzstyle{input neuron}=[neuron, fill=green!80]; %
    \tikzstyle{output neuron}=[neuron, fill=red!80]; %
    \tikzstyle{hidden neuron}=[neuron, fill=cyan];
    \tikzstyle{annot} = [text width=3cm, text centered]
    \tikzstyle{alabel} = [pos=0.9, above=0.5, color=black]
    \tikzstyle{blabel} = [pos=0.1, below=0.5, color=black]

    \foreach \name / \y in {1,...,\insize}
        \node[input neuron] (I_\name) at (0,-\y) {}; 

    \foreach \name / \y in {1,...,\insize}
        \path[yshift=0cm]
        node[hidden neuron] (H_\name) at (\layersep,-\y cm) {};

    \foreach \name / \y in {1,...,\outsize}
        \path[yshift=.5cm]
        node[output neuron] (O_\name) at (2*\layersep,-\y cm) {};

    \foreach \source in {1,...,\insize} {
      \draw[-latex, ultra thick, color=gray] (I_\source) -- node[alabel] {$+$} (H_\source);
    }

    \draw[latex-latex, ultra thick, color=gray] (H_1) -- node[alabel] {$+$} node[blabel] {$-$} (O_4);
    \draw[latex-latex, ultra thick, color=gray] (H_3) -- node[blabel, pos=0.9] {$+$} node[blabel] {$-$} node[pos=0.3, sloped, above=0.1, color=black] {$A_{ij}$} (O_4);
    \draw[latex-latex, ultra thick, color=gray] (H_6) -- node[blabel, pos=0.9] {$+$} node[blabel] {$-$} (O_4);

    \foreach \source in {2, 4, 5} {
      \draw[-, ultra thick, dotted, color=gray] (H_\source) -- (O_4);
    }

    \draw[-latex, ultra thick, color=gray] (O_4) -- node[pos=1.2, color=black] {$x_j$} (2*\layersep+1.2cm, -3.5);

    \node[annot, above of=H_1, node distance=1.5cm] (hl) {$r$};
    \node[left of=hl] {$y$};
    \node[annot,right of=hl] {$x=\thres(A^Ty)$};
    \node[annot, right of=O_7, node distance=2cm] {Hebbian rule \\$\nabla A_{ij} = \eta r_ix_j$};

\end{tikzpicture}
